\documentclass[10pt,twocolumn,letterpaper]{article}

\usepackage[pagenumbers]{cvpr} 

\usepackage{multirow}
\usepackage{makecell}
\usepackage[table]{xcolor}
\usepackage{amsthm}
\usepackage{algorithm}
\usepackage{algpseudocode}
\usepackage{stfloats}
\usepackage{titletoc}
\usepackage{footmisc}

\algrenewcommand\algorithmicrequire{\textbf{Input:}}
\algrenewcommand\algorithmicensure{\textbf{Output:}}

\usepackage{aliascnt}
\theoremstyle{plain}
\newtheorem{theorem}{Theorem}[section]
\newaliascnt{lemma}{theorem}
\newtheorem{lemma}[lemma]{Lemma}
\aliascntresetthe{lemma}
\newaliascnt{proposition}{theorem}

\aliascntresetthe{proposition}
\newaliascnt{corollary}{theorem}

\aliascntresetthe{corollary}

\theoremstyle{definition}
\newaliascnt{definition}{theorem}
\newtheorem{definition}[definition]{Definition}
\aliascntresetthe{definition}
\newaliascnt{assumption}{theorem}
\newtheorem{assumption}[assumption]{Assumption}
\aliascntresetthe{assumption}
\AtEndPreamble{%
  \crefname{assumption}{Assumption}{Assumptions}%
  \Crefname{assumption}{Assumption}{Assumptions}}

\newtheorem*{theorem*}{Theorem}
\newtheorem*{lemma*}{Lemma}
\newtheorem*{proposition*}{Proposition}
\newtheorem*{corollary*}{Corollary}

\theoremstyle{definition}
\newtheorem*{definition*}{Definition}
\newtheorem*{assumption*}{Assumption}

\newcommand{\dpos}[1]{{\scriptsize\textcolor{cvprblue}{$+$#1}}}
\newcommand{\dneg}[1]{{\scriptsize\textcolor{cvprblue}{$-$#1}}}

\definecolor{cvprblue}{rgb}{0.21,0.49,0.74}
\usepackage[pagebackref,breaklinks,colorlinks,allcolors=cvprblue]{hyperref}

\def\confName{CVPR}
\def\confYear{2027}

\title{WAMotor: Accelerating World Action Models via Speculative Inference under Dual Nonuniform Denoising Dynamics}

\author{
Zexun Lin\textsuperscript{1}\textsuperscript{2}\quad
Yuan Feng\textsuperscript{1}\textsuperscript{2}\quad
Xike Xie\textsuperscript{1}\textsuperscript{2}\thanks{Corresponding Author.}\\
\textsuperscript{1}University of Science and Technology of China (USTC)\\
\textsuperscript{2}Data Darkness Lab, MIRACLE Center, Suzhou Institute for Advanced Research, USTC\\
{\tt\small \{zexunlin, yfung\}@mail.ustc.edu.cn, xkxie@ustc.edu.cn}
}

\begin{document}
\maketitle
\begin{abstract}
    World Action Models (WAMs) unify world modeling and action generation for robotic control by integrating flow-matching velocities, yet suffer from high multi-step denoising latency. Recent studies have explored speculative inference to accelerate WAMs by drafting and verifying denoising steps in parallel. However, existing methods either require training task-specific draft models or overlook critical WAM denoising dynamics when reusing past velocities as drafts. We revisit the denoising dynamics of WAMs and identify a dual nonuniformity that breaks standard velocity-reuse assumptions: velocity smoothness is nonuniform as adjacent velocity residuals differ sharply at early and final steps, so reuse there yields large residuals, which nonuniform integration time steps further translate into local state errors that accumulate into the final action distribution. Accordingly, we formalize this by theoretically deriving a trajectory error tolerance bound that characterizes how velocity residuals and integration time steps affect the final action distribution. Based on this, we propose WAMotor, the first speculative inference method tailored to the dual nonuniformity in WAMs. It selects reliable anchor velocities by deferring speculation and employs time-step-aware verification that scales acceptance thresholds by integration time step size. WAMotor offers plug-and-play benefits, enabling seamless integration with prior WAMs. Extensive evaluations on LIBERO, RoboTwin 2.0, and LIBERO-Plus show that WAMotor achieves 2.86$\times$--4.28$\times$ end-to-end speedups while maintaining task success rates. Our code is available at: \url{https://github.com/Overdue-Lin/WAMotor}
\end{abstract}
    
\section{Introduction}

World Action Models (WAMs) offer a new paradigm for embodied policy learning by integrating visual dynamics modeling with action generation, providing predictive representations for robotic control~\cite{gr1-wu,dreamzero-ye,lingbotva-li,lingbotva2-zhang,fastwam-yuan,fasterwam-zhao,cosmos-kim,motus2-bi,zerowam-zhou}. 
However, most existing WAMs generate actions via flow matching~\cite{fm-lipman, rectified-liu}, which uses predicted velocities to update noisy actions over multiple Euler steps, incurring substantial inference cost and limiting task efficiency.
Speculative inference~\cite{sd-leviathan} provides a promising solution for accelerating flow-matching-based WAMs. It achieves this by generating draft denoising trajectories and leveraging parallel verification to advance multiple denoising steps in a single round, reducing inference latency while maintaining outputs close to those of the target model.

\begin{figure}
    \centering
    \includegraphics[width=1\linewidth]{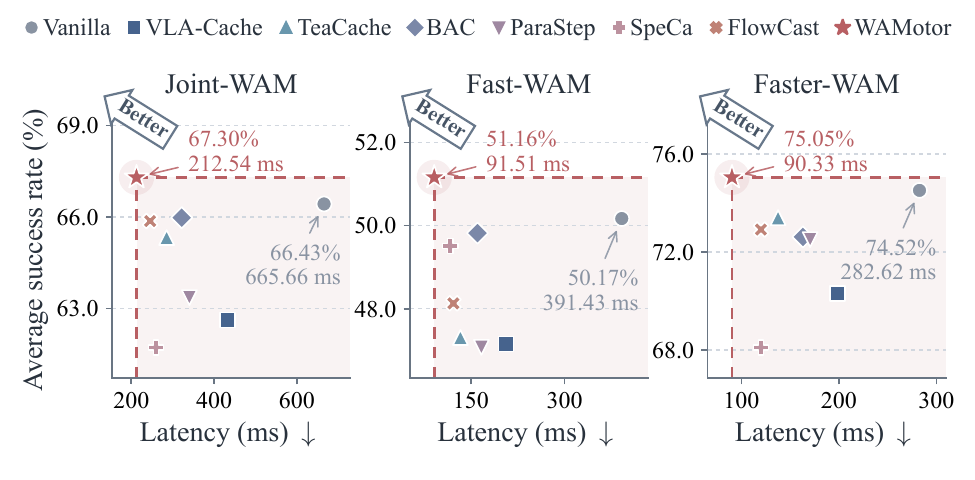}
    \caption{Performance and efficiency gains by WAMotor across different WAMs on LIBERO-Plus. WAMotor achieves pareto optimality (see experiments for other benchmarks).}
    \label{fig:libero-plus-pareto}
\end{figure}

\begin{figure*}
    \centering
    \begin{minipage}[c]{0.45\textwidth}
        \begin{subfigure}{\textwidth}
            \centering
            \includegraphics[width=\textwidth]{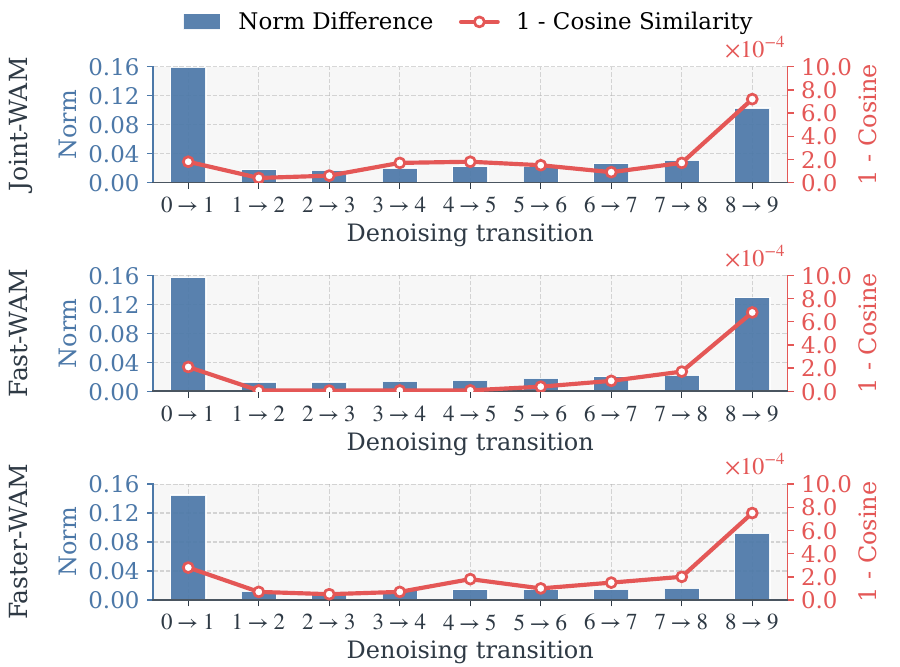}
            \caption{Nonuniform Velocity Smoothness across three WAMs}
            \label{fig:motivation-left}
        \end{subfigure}
    \end{minipage}
    \hfill
    \begin{minipage}[c]{0.54\textwidth}
        \begin{subfigure}{\textwidth}
            \centering
            \includegraphics[width=\textwidth]{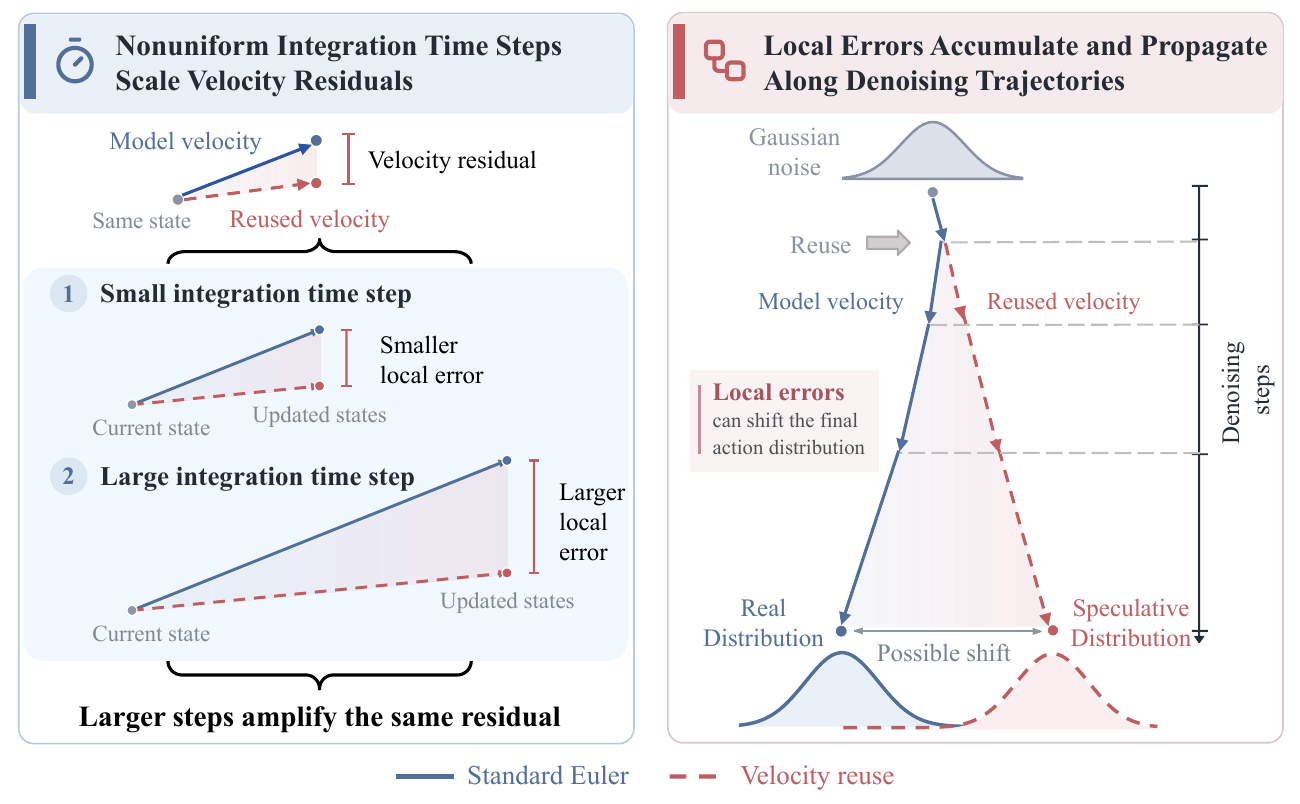}
            \caption{From Velocity Residuals to Action Distribution Errors}
            \label{fig:motivation-middle-sub4}
        \end{subfigure}
    \end{minipage}
    \caption{Motivating observations for WAMotor. 
    \subref{fig:motivation-left} shows differences in magnitude and direction between adjacent velocity predictions, with $i \to i+1$ denoting the compared denoising steps.
    \subref{fig:motivation-middle-sub4} illustrates how nonuniform integration time steps scale velocity residuals induced by velocity reuse into different local errors, which can accumulate along the trajectory and shift the final action distribution.}
    \label{fig:motivation-overall}
\end{figure*}%

Existing speculative inference methods for embodied policies~\cite{waspecdec-wen,kerv-zheng,heisd-zheng,specvla-wang,realtimevlaflash-niu} rely on lightweight draft models trained for specific tasks, which must be retrained whenever the robot or task changes, introducing additional training overhead.
In parallel, recent work on image and video generation exploits velocity smoothness in flow matching by reusing a previously computed velocity as an anchor to approximate subsequent ones~\cite{flowcast-jyoti,parastep-wang}. 
This allows us to construct readily available drafts for speculative inference without additional training.
However, as shown in \cref{fig:motivation-overall}, we revisit the denoising dynamics of WAMs and uncover a previously overlooked dual nonuniformity that breaks the smoothness assumption underlying prior velocity reuse. 
First, velocity smoothness is nonuniform along WAM denoising trajectories, with pronounced changes between adjacent velocities at the beginning and end of denoising. 
Second, nonuniform integration time steps translate comparable velocity residuals into different local state errors, which accumulate and propagate to the final action distribution. 
These observations make it necessary for us to fundamentally rethink and naturally reshape two key questions in WAM’s speculative inference:

{
    \centering
    \colorbox{lightgray!20}{\parbox{0.98\linewidth}{
        \begin{enumerate}
            \item \textbf{\textit{When to Draft:}} \textit{Which velocities in WAMs can serve as reliable anchors for reuse?}
            \item \textbf{\textit{How to Verify:}} \textit{What factors affect the final action distribution error induced by velocity reuse in WAMs?}
        \end{enumerate}
    }}
}

To answer these questions, we formulate speculative inference as a trajectory error tolerance problem. Concretely, we bound the discrepancy between speculative outputs and reference Euler outputs using a weighted sum of anchor velocity reuse residuals. Our analysis shows that unreliable anchors introduce larger velocity residuals, increasing the error bound as these residuals propagate through subsequent nonuniform integration steps and affect the final action distribution. Based on this analysis, we propose \textit{WAMotor}, the first speculative inference method tailored to this dual nonuniformity in WAMs. It is built on two mechanisms: First, we perform \textit{Reliable Anchor Velocity Selection} by deferring drafting until after the first Euler update. Second, we design a \textit{Time-Step-Aware Verification} rule based on the quadratic relationship between squared velocity residuals and integration time step size, tightening verification for larger steps and relaxing it for smaller ones while providing a trajectory error tolerance guarantee.

We seamlessly integrate WAMotor into three representative WAMs, Joint-WAM~\cite{fastwam-yuan} (a controlled implementation representing Joint-WAMs~\cite{fasterwam-zhao,dreamwam-yuan,4dwam-yang}), Fast-WAM~\cite{fastwam-yuan} and Faster-WAM~\cite{fasterwam-zhao}, replacing their standard flow matching denoising pipelines. Comprehensive evaluations on LIBERO, RoboTwin 2.0, and LIBERO-Plus demonstrate that our method achieves greater speedups with near-lossless task performance relative to vanilla models, outperforming cache-based and other speculative inference baselines. Our experiments show that WAMotor accelerates flow matching inference by 3.24$\times$--6.62$\times$, yielding overall speedups of 2.86$\times$--4.28$\times$ over vanilla models. Our main contributions are summarized as follows.

\begin{itemize}
    \item In this paper, we revisit WAM denoising dynamics and uncover a previously overlooked dual nonuniformity in velocity smoothness and integration time steps.
    
    \item We propose WAMotor, a training-free speculative inference method combining reliable anchor selection with time-step-aware verification. It achieves 2.86$\times$--4.28$\times$ end-to-end speedups across three WAMs while maintaining comparable task success rates.

    \item We further derive trajectory error bounds linking velocity-reuse residuals and integration time steps to final action distribution error. This analysis provides a principled foundation for future WAM acceleration, paving the way for more efficient and reliable embodied control.
\end{itemize}

\section{Related Work}

\subsection{Efficient World Action Models}

World Action Models predict future world states, providing policies with temporal information beyond the current observation~\cite{dreamzero-ye,lingbotva-li,lingbotva2-zhang,fastwam-yuan,vpp-hu,vpp2-guo}, but their video and action denoising incurs substantial inference overhead.
Models~\cite{gigaworld-ye,selfwam-pan,internw0delta-miao} exemplified by Fast-WAM~\cite{fastwam-yuan} argue that the benefits of future modeling arise primarily during training, making explicit future video prediction optional at inference time. 
However, Faster-WAM~\cite{fasterwam-zhao} and related approaches~\cite{efficientwam-li,simplewam-zhou} show that removing inference-time future modeling can compromise robustness in out-of-distribution (OOD) settings. They retain future information as a conditioning signal for action generation while reducing its computational cost through mechanisms such as KV-Fusion and sparse video--action interaction.
Our work is complementary to this line of research, exploiting the velocity smoothness characteristics of WAMs to accelerate action generation through velocity reuse.


\subsection{Speculative Inference for Embodied Policies}

Speculative inference reduces sequential model evaluations by generating candidate outputs and verifying them in parallel. It has been widely studied in large language models~\cite{sd-leviathan,ss-chen,eagle-li,eagle2-li,eagle3-li,dflash-chen}. 
Recent methods~\cite{kerv-zheng,heisd-zheng,waspecdec-wen} such as SpecVLA~\cite{specvla-wang} and FLASH~\cite{realtimevlaflash-niu} extend speculative inference to embodied policies by training lightweight draft models. 
However, these methods require separate draft models to be trained for different task domains, introducing additional training overhead and complicating their transfer to WAMs.
Alternatively, FlowCast~\cite{flowcast-jyoti} exploits trajectory smoothness in flow matching to construct speculative drafts by reusing velocity predictions without additional training.
However, such velocity-reuse methods implicitly assume uniform trajectory smoothness and overlook how integration time steps scale reuse errors into the final action distribution. 
In contrast, WAMotor addresses the dual nonuniformity by deferring drafting past abrupt velocity transitions and weighting verification by integration time step size\footnote{We discuss caching methods that are orthogonal to ours in \cref{sec:addition-related-work}.}.

\section{Observation and Motivation}
\label{sec:motivation}

\subsection{Not Every Velocity Is Suitable to Be an Anchor}
\label{sec:motivation-1}

Recent studies on speculative inference for flow matching construct draft trajectories by approximating the velocity field.
To characterize the variation of velocities in WAMs, we analyze differences between consecutive velocity predictions along the denoising trajectories of several WAMs. 
As shown in \cref{fig:motivation-left}, adjacent velocities remain close in both direction and magnitude throughout most intermediate denoising steps. This local smoothness provides a natural draft for predicting the subsequent velocity field, enabling velocity reuse without an auxiliary draft model. 

However, this local smoothness breaks down near the beginning and end of denoising, where consecutive velocities change sharply in direction and magnitude. 
These changes limit how well the preceding velocity represents subsequent updates. Speculation based on such anchor velocity may face early rejection or, even when the candidate is ultimately accepted, result in substantial deviations in the trajectory\footnote{We provide further visualization and quantitative analysis in \cref{sec:heatmap-visualization} and \cref{sec:app-velocity-stats}, respectively.}. 
Given this nonuniform velocity smoothness, WAMotor defers speculation until after the first standard Euler update, using the velocity at the updated state as the initial anchor.

\subsection{Velocity Residual Alone Does Not Determine Final Action Distribution Deviation}
\label{sec:motivation-2}

Having identified a more reliable anchor velocity for speculation, the second question is which drafted velocities should be accepted. Existing speculative inference methods for flow matching assess the residual between the reused velocity and the model prediction at each candidate state using a velocity-space MSE criterion.

However, flow matching samplers often employ nonuniform integration time step sizes across denoising steps~\cite{srft-esser,iddpm-quinn,edm-karras,dj-chen}. As shown in \cref{fig:motivation-middle-sub4}, under such schedules, the same velocity discrepancy can have different effects on the action state update depending on the corresponding integration interval. Moreover, deviations introduced by accepted updates may accumulate and propagate along the denoising trajectory. A uniform velocity-MSE threshold therefore does not provide uniform control over the velocity residuals introduced by speculation. In the following analysis, WAMotor establishes a link between velocity residuals, integration time steps, and final action distribution, motivating a time-step-aware verification criterion.

\section{Methodology}

\subsection{Preliminaries}

\paragraph{Flow Matching.}

Most existing embodied policies generate actions via flow matching~\cite{fm-lipman,act-zhao,diffusion-policy,pi0-black,lingbotva-li,dreamwam-yuan}. 
Specifically, let $a_k\in\mathbb{R}^{H\times D}$ denote normalized actions conditioned on model-specific context $c_k$, including observations, language, proprioception, and any future representations. 
Starting from Gaussian noise $a_0=z\sim\mathcal{N}(0,I)$, the learned velocity field $v_\theta$ progressively transforms the noise into actions through $N$ Euler steps: $a_{k+1}=a_k+\Delta t_k\,v_\theta(a_k,t_k;c_k)$ for $k=0,\ldots,N-1$, where $H$ and $D$ denote the action horizon and dimension, respectively, and $\Delta t_k=t_{k+1}-t_k$ is the integration time step.

\paragraph{Speculative Inference.}

To accelerate action generation with controlled approximation error,  speculative inference constructs a speculative trajectory $\{a_k^s\}_{k=0}^{N}$ initialized at $a_0^s=a_0$. At each anchor step $m$, the current speculative state $a_m^s$ and its anchor velocity $u_m=v_\theta(a_m^s,t_m;c_m^s)$ are leveraged to draft $q_m=\min(w,N-1-m)$ candidates, where $w$ denotes the speculation window cap. Specifically, the $j$-th candidate is generated as

\begin{equation}
	\label{eq:flow-speculative}
	\hat{a}_{m,j} = a_m^s + u_m\sum_{i=0}^{j-1}\Delta t_{m+i}, \, j=1\cdots q_m
\end{equation}

To verify these drafts, a batched model call evaluates $\hat{v}_{m,j}=v_\theta(\hat{a}_{m,j},t_{m+j};\hat c_{m,j})$ for all candidates, yielding velocity residuals $r_{m,j}=u_m-\hat{v}_{m,j}$. Candidates are verified against a predefined criterion, and the longest accepted prefix, of length $\ell$, is committed as $a_{m+j}^s=\hat{a}_{m,j}$ for $j=1,\ldots,\ell$.
For WAMs with evolving future conditioning, speculative conditioning $\hat c_{m,j}$ is likewise constructed, and $c_{m+j}^s$ is updated accordingly for retained candidates.

\subsection{Problem Formalization: Revisiting Denoising Dynamics via Trajectory Error Tolerance}

Recent work has demonstrated that the smoothness of flow matching velocity fields can be exploited to construct speculative trajectories by reusing previously evaluated velocities~\cite{parastep-wang,flowcast-jyoti}. 
However, as observed in \cref{sec:motivation}, the reliability of an anchor velocity varies across denoising stages, while nonuniform integration time steps translate comparable velocity residuals into different significantly local state errors. 
The central challenge is therefore to identify reliable anchors and devise an appropriate verification strategy so that speculation accelerates inference while keeping the speculative action distribution close to the full Euler output distribution. 
To formalize this objective, we define a Trajectory $\delta$-error Tolerance that bounds the discrepancy.

\begin{definition}[Trajectory $\delta$-error Tolerance]
	A speculative strategy satisfies a Trajectory $\delta$-error Tolerance if
	
	\begin{equation}
		\label{eq:terminal-discrepancy}
		\|a_N^s-a_N\|_{\mathcal{A}} \le \delta
	\end{equation}
	
	Here, $\| \cdot \|_{\mathcal{A}}^2$ denotes maximum per-token MSE over the blocks, which ensures the acceptance criterion applies to every token.
\end{definition}

\subsection{Stage I: Reliable Anchor Velocity Selection}
\label{sec:anchor}

As observed in \cref{sec:motivation-1}, not every denoising step provides an equally reliable starting point for speculation. To quantify how anchor velocity choice affects the Trajectory $\delta$-error Tolerance, we adopt the standard Lipschitz continuity assumption~\cite{dpm-solver-lu,ebfm-benton,flowcast-jyoti,vla-tyt-li}, which has been widely used to analyze the stability of flow matching and diffusion models.

\begin{assumption}[Lipschitz Continuity]
\label{asm:lipschitz}
    At each denoising step $k$, the difference between predicted velocities is bounded by a weighted sum of differences in action states $x$ and $y$ and conditioning inputs $c$ and $c'$, with constants $L_k,C_k\ge0$:
	
	\begin{equation}
		\label{eq:lipschitz}
        \begin{aligned}
            &\|v_\theta(x, t_k; c) - v_\theta(y, t_k; c')\|_{\mathcal{A}} \le \\
            &\quad L_k \|x - y\|_{\mathcal{A}} + C_k\|c-c'\|_{\mathcal A}.
        \end{aligned}
	\end{equation}
\end{assumption}

A deviation between the speculative and reference trajectories at step $k$ can be amplified through subsequent denoising dynamics. Let $h_k = |\Delta t_k|$ and define the propagation factor $G_k = \prod_{j=k+1}^{N-1}(1 + h_j L_j)$ and the propagated conditioning-error contribution $B_{\mathrm{cond}}=\sum_{k=0}^{N-1}G_kh_kC_k\|c_k^s-c_k\|_{\mathcal A}$, where $c_k^s$ and $c_k$ denote the conditioning inputs along the speculative and reference trajectories, respectively. 
In \cref{thm:traj}, we bound the trajectory error tolerance in terms of the per-step velocity residuals accepted into the speculative trajectory.

\begin{theorem}
	\label{thm:traj}

    Trajectory $\delta$-error tolerance can be given by a weighted sum of anchor velocity reuse residuals plus the propagated conditioning-error contribution, where larger residuals increase the bound with other terms held fixed, and earlier residuals carry no smaller propagation factors.
	
	\begin{equation}
	    \label{eq:traj-bound}
        \begin{aligned}
            &\|a_N^s - a_N\|_{\mathcal{A}} \le \delta \\ := & \sum_{k \in \mathcal{I}} G_k\, h_k\, \|r_{M_k,k-M_k}\|_{\mathcal{A}} + B_{\mathrm{cond}},
        \end{aligned}
	\end{equation}
	
    where $\mathcal I$ denotes the set of update indices at which an earlier anchor velocity is reused in the transition from $k$ to $k+1$. For each $k\in\mathcal I$, $M_k<k$ denotes the index of the anchor whose velocity is reused.
\end{theorem}

\begin{proof}
	See \cref{sec:proof-thm-traj} for details.
\end{proof}

\cref{thm:traj} has a direct implication for anchor selection. An anchor whose velocity $u_m$ poorly represents the velocity field at subsequent draft states $\hat{a}_{m,j}$ produces large residuals $\|r_{m,j}\|_{\mathcal{A}}$, driving up the bound proportionally. 

As established in \cref{sec:motivation-1}, the velocity field of WAMs is not uniformly smooth, with pronounced changes in both velocity direction and magnitude between the first two denoising steps. As characterized by \cref{eq:traj-bound}, the larger residuals from reusing the initial velocity $u_0$ enlarge the final error bound through propagation across subsequent updates. Motivated by this, WAMotor employs \textit{Reliable Anchor Velocity Selection}, which conducts one standard Euler update before the first speculative window and uses $u_1$ at $a_1^s$ as the initial anchor, while velocity differences between the final two steps are handled by verification and correction.

\subsection{Stage II: Time-Step-Aware Verification}
\label{sec:threshold}

Given a reliable anchor velocity, the remaining question is how to reliably verify velocity reuse in parallel at each denoising step. Existing speculative inference methods for flow matching typically assess draft accuracy in velocity space, checking whether the reused velocity approximately matches the model prediction. However, \cref{thm:traj} shows that the trajectory error depends on both the velocity residual and integration time step size. The following lemma makes this dependence explicit for a single update.

\begin{lemma}
	\label{lem:local}
	The MSE between reusing the anchor velocity and performing an exact Euler update from the same speculative state scales quadratically with the integration time step size for a fixed velocity residual.
	\begin{equation}
		\label{eq:local-error}
		\begin{aligned}
			&\left\|
			(a_k^s+\Delta t_{k} u_{M_k})
			-(a_k^s+\Delta t_{k}\hat{v}_{M_k,k-M_k})
			\right\|_{\mathcal{A}}^2\\
			&\qquad=h_{k}^2\|r_{M_k,k-M_k}\|_{\mathcal{A}}^2.
		\end{aligned}
	\end{equation}
\end{lemma}

\begin{proof}
	See \cref{sec:proof-lem-local} for details.
\end{proof}

\cref{lem:local} shows that the local state error depends on both the velocity residual and the integration time step size. Even a small velocity residual can incur a substantial state error when amplified by a large integration time step. For a fixed velocity discrepancy, the local state MSE scales quadratically with the integration time step size. 

\paragraph{Verification.} Motivated by the time step-size dependence established in \cref{lem:local}, we propose \textit{Time-Step-Aware Verification} which weights squared velocity-reuse residuals by squared integration time step sizes under nonuniform schedules, thereby further tightening the trajectory error tolerance. Let $h^{\mathrm{mid}}$ denote the median integration time step size. We tighten velocity verification for steps larger than $h^{\mathrm{mid}}$ and relax it for smaller steps, enforcing a uniform local state-error tolerance by accepting reuse if\footnote{We empirically validate and analyze the choice of $h^{\mathrm{mid}}$ in \cref{sec:sensitivity-ref}.}

\begin{equation}
	\label{eq:time-normalized-verification}
	\|r_{M_k,k-M_k}\|_{\mathcal A}^2
	\left(\frac{h_k}{h^{\mathrm{mid}}}\right)^2
	\le\varepsilon.
\end{equation}

The verification threshold $\varepsilon\ge0$ controls the local state errors permitted during velocity reuse. The following theorem relates this threshold to the trajectory error tolerance.

\begin{theorem}
	\label{thm:time-bound}
	
	Time-Step-Aware Verification with the median reference step $h^{\mathrm{mid}}$ and threshold $\varepsilon\ge0$ guarantees trajectory $\delta_{\mathrm{time}}$-error tolerance.
	
	\begin{equation}
		\label{eq:time-bound}
		\|a_N^s - a_N\|_{\mathcal{A}} \;\le\;\delta_{\mathrm{time}}=\; h^{\mathrm{mid}}\sqrt{\varepsilon} \sum_{k\in\mathcal I}G_k + B_{\mathrm{cond}}
	\end{equation}
\end{theorem}

\begin{proof}
	See \cref{sec:proof-thm-time-bound} for details.
\end{proof}

\cref{thm:time-bound} establishes a theoretical link between the verification threshold $\varepsilon$ and the trajectory error tolerance $\delta_{\mathrm{time}}$, providing a basis for controlling the final action distribution discrepancy through threshold selection.

\subsection{Integration into Existing World Action Models}

Combining Reliable Anchor Velocity Selection with Time-Step-Aware Verification yields WAMotor (\cref{algorithm}), which replaces the original inference sampler without an auxiliary draft model and accelerates action generation via velocity-reuse drafting and parallel verification.

\paragraph{Model-Specific Integration.} We demonstrate the plug-and-play compatibility of WAMotor across three representative WAMs by seamlessly integrating it into their existing action generation pipelines. It supports joint video--action extrapolation in Joint-WAM\footnote{See \cref{sec:pseudocode-joint} for the pseudocode with joint video--action prediction.}, action-only speculation with a fixed observed-frame visual cache in Fast-WAM, and speculation with cached future context and sparse video--action interaction in Faster-WAM.

\begin{algorithm}
    \caption{WAMotor}
    \label{algorithm}
    \begin{algorithmic}[1]
        \Require Gaussian noise $a_0^s$, Diffusion step $N$
		\Ensure Action $a_N^s$
        \State $a_1^s \gets a_0^s + \Delta t_0\,v_\theta \left(a_0^s,t_0;c_0^s\right)$ \Comment{Stage I}
        \State $m \gets 1$, $u_m=v_\theta(a_m^s,t_m;c_m^s)$
        \While{$m < N - 1$}
            \State $\hat a_{m,0}=a_m^s$, $\hat v_{m,0}=u_m$
            \State \textbf{Draft:} $\hat{a}_{m,j} = a_m^s + u_m\sum\limits_{i=0}^{j-1}\Delta t_{m+i}$,\, $1 \le j \le q_m$
            \State \textbf{Verify:} $\hat{v}_{m,j} \gets v_\theta\left(\hat{a}_{m,j},t_{m+j};\hat c_{m,j}\right)$, $e_{m+j} \gets \|u_m-\hat v_{m,j}\|_{\mathcal A}^2\cdot\left(\frac{h_{m+j}}{h^{\mathrm{mid}}}\right)^2$, $1 \le j \le q_m$ \Comment{Stage II}
            \State Find first index $i \in [1,q_m]$ such that $e_{m+i} > \varepsilon$
            \If{$e_{m+j} \le \varepsilon$, $1 \le j \le q_m$}
                \State $a_{m+j}^s \gets \hat{a}_{m,j}$, $1 \le j \le q_m$
                \State $u_{m+q_m} \gets \hat{v}_{m,q_m}$, $m \gets m + q_m$
            \Else
                \State \textbf{Correction:} $a_{m+j}^s \gets \hat{a}_{m,j}$, $1 \le j < i$
                \State $a_{m+i}^s\gets \hat a_{m,i-1}+\Delta t_{m+i-1}\hat v_{m,i-1}$
                \State $u_{m+i}\gets v_\theta(a_{m+i}^s,t_{m+i};c_{m+i}^s)$, $m \gets m + i$
            \EndIf
        \EndWhile
        \State $a_N^s \gets a_{N-1}^s+\Delta t_{N-1}\,u_{N-1}$
        \State \Return $a_N^s$
    \end{algorithmic}
\end{algorithm}

\begin{table*}[t]
    \centering

    \begin{minipage}[t]{0.48\textwidth}
        \centering
        \small
        \setlength{\tabcolsep}{1.3pt}
        \caption{Success rates (\%) on LIBERO. Bold and underlined numbers denote the best and second-best results among all methods except Vanilla, respectively.}
        \label{tab:libero}
        \begin{tabular}{clcccccc}
        \toprule
        Model & Baseline & Spa. & Obj. & Goa. & Lon. & Avg. $\uparrow$ & Lat. (ms) $\downarrow$ \\
        \midrule
        
        \multirow{8}{*}{\rotatebox[origin=c]{90}{Joint-WAM}}
        & Vanilla
        & 99.40 & 98.80 & 98.60 & 96.60 & 98.35 & 660.83 \\
        & VLA-Cache
        & \underline{98.80} & 98.60 & 96.80 & 94.60 & 97.20 & 434.68 \\
        & TeaCache
        & 98.20 & \textbf{99.00} & 96.80 & 95.00 & 97.25 & 276.87 \\
        & BAC
        & \underline{98.80} & \underline{98.80} & \underline{97.40} & \underline{95.80} & \underline{97.70} & 331.91 \\
        & ParaStep
        & 97.40 & 97.60 & 95.80 & 94.00 & 96.20 & 346.39 \\
        & SpeCa
        & 97.20 & 98.20 & 95.20 & 92.40 & 95.75 & \underline{245.89} \\
        & FlowCast
        & \underline{98.80} & 98.60 & 97.20 & \underline{95.80} & 97.60 & 246.79 \\
        & \cellcolor{cvprblue!20}w/ ours
        & \cellcolor{cvprblue!20}\textbf{99.00}
        & \cellcolor{cvprblue!20}\underline{98.80}
        & \cellcolor{cvprblue!20}\textbf{97.80}
        & \cellcolor{cvprblue!20}\textbf{96.40}
        & \cellcolor{cvprblue!20}\textbf{98.00}
        & \cellcolor{cvprblue!20}\textbf{216.44} \\
        
        \midrule
        
        \multirow{8}{*}{\rotatebox[origin=c]{90}{Fast-WAM}}
        & Vanilla
        & 97.20 & 99.80 & 96.40 & 95.00 & 97.10 & 387.54 \\
        & VLA-Cache
		& \underline{97.60} & 99.20 & 95.60 & 93.60 & 96.50 & 203.43 \\
		& TeaCache
		& 97.40 & 99.20 & \underline{95.80} & 94.20 & 96.65 & 137.78 \\
		& BAC
		& \underline{97.60} & \underline{99.40} & \textbf{96.40} & 94.60 & \underline{97.00} & 161.10 \\
		& ParaStep
		& 96.20 & 98.40 & 94.60 & 92.80 & 95.50 & 169.63 \\
		& SpeCa
		& 94.80 & 98.00 & 93.00 & 90.40 & 94.05 & \underline{109.58} \\
		& FlowCast
		& \underline{97.60} & \underline{99.40} & 95.60 & \underline{95.00} & 96.90 & 124.07 \\
		& \cellcolor{cvprblue!20}w/ ours
		& \cellcolor{cvprblue!20}\textbf{97.80}
		& \cellcolor{cvprblue!20}\textbf{99.60}
		& \cellcolor{cvprblue!20}\underline{95.80}
		& \cellcolor{cvprblue!20}\textbf{95.20}
		& \cellcolor{cvprblue!20}\textbf{97.10}
		& \cellcolor{cvprblue!20}\textbf{92.57} \\
        
        \midrule
        
        \multirow{8}{*}{\rotatebox[origin=c]{90}{Faster-WAM}}
        & Vanilla
        & 98.60 & 99.80 & 98.20 & 97.80 & 98.60 & 275.85 \\
        & VLA-Cache
		& 97.40 & 99.20 & 96.00 & 96.40 & 97.25 & 193.24 \\
		& TeaCache
		& 97.40 & 99.20 & 96.60 & 96.80 & 97.50 & 134.21 \\
		& BAC
		& \underline{98.80} & \textbf{100.00} & \underline{98.40} & \underline{97.80} & \underline{98.75} & 163.10 \\
		& ParaStep
		& 97.40 & 98.40 & 96.60 & 95.80 & 97.05 & 170.81 \\
		& SpeCa
		& 95.60 & 97.20 & 95.20 & 94.20 & 95.55 & \underline{108.57} \\
		& FlowCast
		& 98.60 & \underline{99.80} & 98.20 & \underline{97.80} & 98.60 & 120.38 \\
		& \cellcolor{cvprblue!20}w/ ours
		& \cellcolor{cvprblue!20}\textbf{99.40}
		& \cellcolor{cvprblue!20}\textbf{100.00}
		& \cellcolor{cvprblue!20}\textbf{99.00}
		& \cellcolor{cvprblue!20}\textbf{98.40}
		& \cellcolor{cvprblue!20}\textbf{99.20}
		& \cellcolor{cvprblue!20}\textbf{90.10} \\
        
        \bottomrule
        \end{tabular}
    \end{minipage}
    \hfill
    \begin{minipage}[t]{0.48\textwidth}
        \centering
        \small
        \setlength{\tabcolsep}{4pt}
        \caption{Success rates (\%) on RoboTwin 2.0. Bold and underlined numbers denote the best and second-best results among all methods except Vanilla, respectively.}
        \label{tab:robotwin}
        \begin{tabular}{clcccc}
        \toprule
        Model & Baseline & Clean & Rand. & Avg. $\uparrow$ & Lat. (ms) $\downarrow$ \\
        \midrule
        
        \multirow{8}{*}{\rotatebox[origin=c]{90}{Joint-WAM}}
        & Vanilla
        & 91.02 & 90.38 & 90.70 & 661.53 \\
        & VLA-Cache
		& 82.12 & 84.80 & 83.46 & 432.05 \\
		& TeaCache
		& 86.28 & 88.32 & 87.30 & 269.92 \\
		& BAC
		& \underline{90.80} & \underline{89.76} & \underline{90.28} & 337.29 \\
		& ParaStep
		& 89.32 & 88.28 & 88.80 & 347.45 \\
		& SpeCa
		& 85.94 & 84.22 & 85.08 & 247.33 \\
		& FlowCast
		& 89.42 & 88.82 & 89.12 & \underline{246.15} \\
		& \cellcolor{cvprblue!20}w/ ours
		& \cellcolor{cvprblue!20}\textbf{91.24}
		& \cellcolor{cvprblue!20}\textbf{90.36}
		& \cellcolor{cvprblue!20}\textbf{90.80}
		& \cellcolor{cvprblue!20}\textbf{217.87} \\
        
        \midrule
        
        \multirow{8}{*}{\rotatebox[origin=c]{90}{Fast-WAM}}
        & Vanilla
        & 91.52 & 91.34 & 91.43 & 398.07 \\
        & VLA-Cache
		& 82.66 & 85.88 & 84.27 & 205.55 \\
		& TeaCache
		& 81.24 & \underline{90.52} & 85.88 & 139.27 \\
		& BAC
		& \underline{91.34} & 89.24 & \underline{90.29} & 160.30 \\
		& ParaStep
		& 90.34 & 89.24 & 89.79 & 167.32 \\
		& SpeCa
		& 86.32 & 85.84 & 86.08 & \underline{113.21} \\
		& FlowCast
		& 89.60 & 89.32 & 89.46 & 130.88 \\
		& \cellcolor{cvprblue!20}w/ ours
		& \cellcolor{cvprblue!20}\textbf{91.62}
		& \cellcolor{cvprblue!20}\textbf{91.28}
		& \cellcolor{cvprblue!20}\textbf{91.45}
		& \cellcolor{cvprblue!20}\textbf{98.95} \\
        
        \midrule
        
        \multirow{8}{*}{\rotatebox[origin=c]{90}{Faster-WAM}}
        & Vanilla
        & 92.52 & 92.38 & 92.45 & 282.09 \\
        & VLA-Cache
		& 80.28 & 84.68 & 82.48 & 196.24 \\
		& TeaCache
		& 87.08 & 88.32 & 87.70 & 136.55 \\
		& BAC
		& \underline{91.10} & \textbf{92.08} & \underline{91.59} & 163.88 \\
		& ParaStep
		& 90.08 & 90.02 & 90.05 & 168.36 \\
		& SpeCa
		& 85.38 & 84.26 & 84.82 & \underline{113.38} \\
		& FlowCast
		& 90.04 & 90.96 & 90.50 & 124.65 \\
		& \cellcolor{cvprblue!20}w/ ours
		& \cellcolor{cvprblue!20}\textbf{92.24}
		& \cellcolor{cvprblue!20}\underline{91.86}
		& \cellcolor{cvprblue!20}\textbf{92.05}
		& \cellcolor{cvprblue!20}\textbf{98.70} \\
        
        \bottomrule
        \end{tabular}
    \end{minipage}

\end{table*}

\begin{table*}[t]
    \centering
    \small
    \setlength{\tabcolsep}{7pt}
    \caption{Success rates (\%) on LIBERO-Plus across seven distribution shifts. Bold and underlined numbers denote the best and second-best results among all methods except Vanilla, respectively.}
    \label{tab:libero_plus}
    \begin{tabular}{clccccccccc}
    \toprule
    Model & Baseline
    & Camera & Robot & Lang. & Light & Backg. & Noise & Layout
    & Avg. $\uparrow$ & Lat. (ms) $\downarrow$ \\
    \midrule
    
    \multirow{8}{*}{\rotatebox[origin=c]{90}{Joint-WAM}}
    & Vanilla
    & 36.44 & 62.69 & 91.39 & 90.74 & 55.15 & 49.03 & 79.55 & 66.43 & 665.66 \\
    & VLA-Cache
	& 34.50 & 58.14 & 85.81 & 87.10 & 46.67 & 48.20 & 78.01 & 62.63 & 432.57 \\
	& TeaCache
	& 35.75 & \underline{62.04} & 88.61 & 89.83 & 52.00 & \underline{49.63} & \underline{79.48} & 65.33 & 285.34 \\
	& BAC
	& \underline{36.45} & 61.24 & \underline{90.54} & 90.30 & \underline{54.53} & 49.61 & 79.21 & \underline{65.98} & 321.58 \\
	& ParaStep
	& 33.51 & 54.43 & 86.12 & 90.23 & 51.09 & 49.41 & 78.75 & 63.36 & 340.32 \\
	& SpeCa
	& 36.23 & 53.24 & 85.36 & 88.26 & 50.27 & 47.22 & 71.56 & 61.73 & 259.47 \\
	& FlowCast
	& 35.74 & 61.92 & 90.37 & \textbf{90.37} & 54.29 & 49.37 & 79.00 & 65.87 & \underline{245.45} \\
	& \cellcolor{cvprblue!20}w/ ours
	& \cellcolor{cvprblue!20}\textbf{36.64}
	& \cellcolor{cvprblue!20}\textbf{64.97}
	& \cellcolor{cvprblue!20}\textbf{90.77}
	& \cellcolor{cvprblue!20}\underline{90.32}
	& \cellcolor{cvprblue!20}\textbf{56.38}
	& \cellcolor{cvprblue!20}\textbf{50.01}
	& \cellcolor{cvprblue!20}\textbf{82.02}
	& \cellcolor{cvprblue!20}\textbf{67.30}
	& \cellcolor{cvprblue!20}\textbf{212.54} \\
    
    \midrule
    
    \multirow{8}{*}{\rotatebox[origin=c]{90}{Fast-WAM}}
    & Vanilla
    & 16.53 & 46.11 & 68.03 & 82.72 & 47.94 & 30.74 & 59.09 & 50.17 & 391.43 \\
    & VLA-Cache
	& 13.99 & 41.04 & 59.22 & 79.85 & 46.99 & 30.55 & 58.41 & 47.15 & 206.15 \\
	& TeaCache
	& 14.14 & 40.15 & 60.28 & 78.95 & 46.66 & \underline{31.67} & 59.34 & 47.31 & 133.38 \\
	& BAC
	& 15.49 & 42.85 & 67.48 & \underline{81.85} & \textbf{48.99} & 30.67 & \textbf{61.41} & \underline{49.82} & 160.71 \\
	& ParaStep
	& 15.98 & 41.54 & 62.74 & 74.29 & 46.33 & 29.30 & 59.33 & 47.07 & 166.93 \\
	& SpeCa
	& \underline{17.11} & \underline{43.91} & \underline{68.82} & \underline{81.85} & 47.53 & 29.56 & 57.76 & 49.51 & \underline{117.00} \\
	& FlowCast
	& 15.15 & 43.47 & 67.42 & 80.33 & 43.98 & 28.42 & 58.12 & 48.13 & 122.19 \\
	& \cellcolor{cvprblue!20}w/ ours
	& \cellcolor{cvprblue!20}\textbf{18.84}
	& \cellcolor{cvprblue!20}\textbf{47.35}
	& \cellcolor{cvprblue!20}\textbf{69.05}
	& \cellcolor{cvprblue!20}\textbf{82.30}
	& \cellcolor{cvprblue!20}\underline{48.65}
	& \cellcolor{cvprblue!20}\textbf{31.71}
	& \cellcolor{cvprblue!20}\underline{60.20}
	& \cellcolor{cvprblue!20}\textbf{51.16}
	& \cellcolor{cvprblue!20}\textbf{91.51} \\
    
    \midrule
    
    \multirow{8}{*}{\rotatebox[origin=c]{90}{Faster-WAM}}
    & Vanilla
    & 52.97 & 71.50 & 95.08 & 96.29 & 61.34 & 62.64 & 81.82 & 74.52 & 282.62 \\
    & VLA-Cache
    & 49.25 & 67.83 & 91.67 & 90.38 & 55.02 & 60.05 & 78.01 & 70.32 & 198.28 \\
    & TeaCache
    & 52.19 & \underline{71.23} & 94.41 & 94.66 & 60.31 & 61.29 & 79.75 & \underline{73.41} & 137.59 \\
    & BAC
    & 51.21 & \textbf{71.64} & 92.53 & 94.08 & 59.64 & 60.29 & 78.97 & 72.62 & 163.14 \\
    & ParaStep
    & 50.89 & 70.35 & 92.21 & \underline{94.85} & \underline{60.32} & \underline{61.37} & 77.60 & 72.51 & 170.55 \\
    & SpeCa
    & 51.88 & 63.16 & 88.15 & 91.66 & 50.33 & 59.95 & 71.62 & 68.11 & \underline{120.03} \\
    & FlowCast
    & \underline{52.98} & 68.86 & \underline{95.08} & 92.82 & 57.51 & 61.28 & \underline{81.88} & 72.92 & 120.07 \\
    & \cellcolor{cvprblue!20}w/ ours
    & \cellcolor{cvprblue!20}\textbf{55.48}
    & \cellcolor{cvprblue!20}69.31
    & \cellcolor{cvprblue!20}\textbf{96.30}
    & \cellcolor{cvprblue!20}\textbf{95.46}
    & \cellcolor{cvprblue!20}\textbf{60.70}
    & \cellcolor{cvprblue!20}\textbf{64.99}
    & \cellcolor{cvprblue!20}\textbf{83.13}
    & \cellcolor{cvprblue!20}\textbf{75.05}
    & \cellcolor{cvprblue!20}\textbf{90.33} \\
    
    \bottomrule
    \end{tabular}
\end{table*}

\section{Experiments}

\subsection{Experimental Setup}

\paragraph{Benchmarks.} We evaluate WAMotor on three simulation benchmarks using task success rate and inference latency. LIBERO~\cite{libero-liu} provides language-conditioned manipulation tasks in four commonly evaluated suites. RoboTwin 2.0~\cite{robotwin2.0-chen} introduces a more challenging bimanual manipulation benchmark comprising 50 tasks, evaluated under clean and domain-randomized settings. LIBERO-Plus~\cite{libero-plus-fei} extends LIBERO to evaluate OOD robustness under seven perturbation categories. We evaluate all 40 LIBERO tasks (50 trials each), 50 RoboTwin 2.0 tasks (100 trials each under both clean and randomized settings), and 10,030 LIBERO-Plus task instances (3 trials each)\footnote{Detailed benchmark information is provided in \cref{sec:detail-bench}.}.

\paragraph{Models.} We evaluate WAMotor on three representative open-source WAMs: Joint-WAM~\cite{fastwam-yuan}, Fast-WAM~\cite{fastwam-yuan}, and Faster-WAM~\cite{fasterwam-zhao}. They span joint video--action denoising, observation-conditioned action generation, and sparse future conditioning, allowing us to assess WAMotor's applicability across architectures.

\paragraph{Baselines.} We use each model’s vanilla inference as the reference. We compare against VLA-Cache~\cite{vla-cache-xu}, TeaCache~\cite{teacache-liu}, and BAC~\cite{bac-ji} as training-free caching baselines, and ParaStep~\cite{parastep-wang} and SpeCa~\cite{speca-liu} as caching baselines inspired by diffusion speculative inference. We also include FlowCast~\cite{flowcast-jyoti} as a flow matching speculative inference baseline that requires no draft model.

\paragraph{Implementation Details.} All experiments are conducted on RTX 5090. Following \cref{algorithm}, we perform one standard Euler update before speculation, set the verification threshold to $\varepsilon=10^{-2}$, and use the total number of denoising steps as the default speculative window cap.

\subsection{Main Results}

\paragraph{LIBERO.} \Cref{tab:libero} summarizes the results across the four LIBERO suites. WAMotor achieves the lowest inference latency across all three WAMs, providing speedups of $3.05\times$--$4.19\times$ over Vanilla  while maintaining comparable task success rates. It also reduces latency by 12.0\%--17.0\% relative to SpeCa, the fastest competing baseline on this benchmark. Meanwhile, WAMotor achieves the highest average success rate among acceleration methods.

\paragraph{RoboTwin 2.0.} \Cref{tab:robotwin} reports results under both clean and randomized evaluation settings. WAMotor achieves speedups of $2.86\times$--$4.02\times$ over Vanilla and reduces latency by 11.5\%--12.9\% relative to the fastest competing baseline for each WAM. Among acceleration methods with reported success rates, WAMotor achieves the highest average success rate on all three WAMs, exceeding BAC, the strongest competing method by this metric, by 0.52, 1.16, and 0.46 percentage points for the three WAMs, respectively. Compared with Vanilla, the average success rates of the three WAMs remain largely comparable.

\paragraph{LIBERO-Plus.} \Cref{tab:libero_plus} evaluates robustness across seven categories of distribution shifts. WAMotor achieves the lowest latency and highest average success rate among acceleration methods for all three WAMs. Relative to Vanilla, it provides speedups of $3.13\times$--$4.28\times$ while increasing average success rates by 0.87, 0.99, and 0.53 percentage points for the three WAMs, respectively. Compared with FlowCast, WAMotor further reduces latency by 13.4\%--25.1\% and improves average success rates by 1.43, 3.03, and 2.13 percentage points, respectively. Across all three WAMs, gains over Vanilla are particularly notable under camera, sensor, and object layout perturbations\footnote{We presents a case study on the accuracy gain in \cref{sec:case-study}.}. These results suggest that WAMotor preserves OOD robustness.

\begin{figure}
	\centering
	\includegraphics[width=1\linewidth]{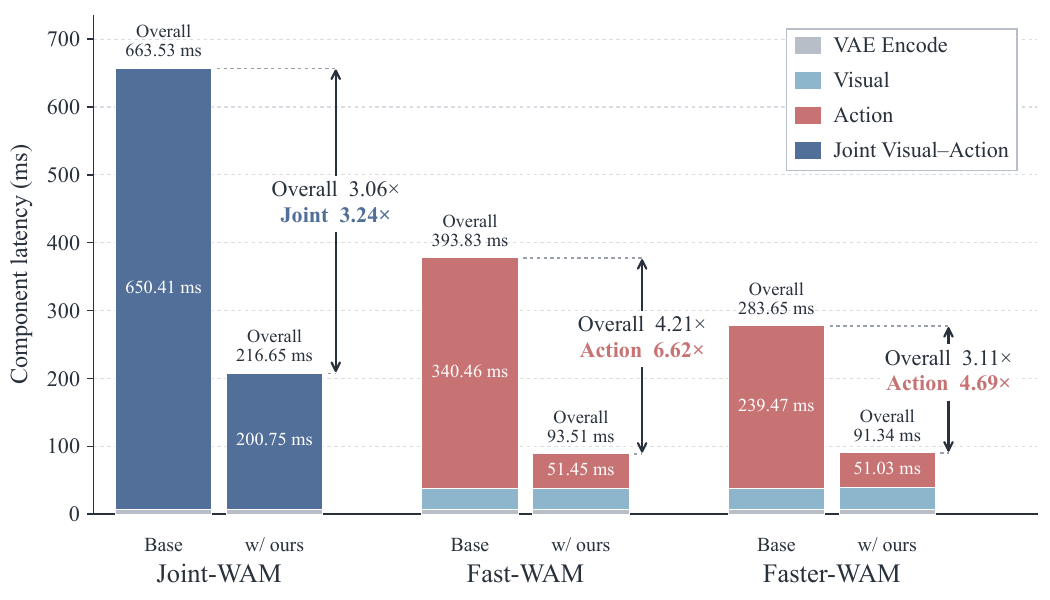}
	\caption{Speedup gains of WAMotor across different WAMs, averaged over 30 runs after 5 warm-up iterations.}
	\label{fig:latency-breakdown}
\end{figure}

\paragraph{Inference Latency Breakdown.} \Cref{fig:latency-breakdown} compares the inference latency of the three WAM variants under their default inference settings and with our method. In this separate experiment, all measurements use LIBERO-like image inputs at a resolution of 224 $\times$ 448 and 10 denoising steps. Our method primarily reduces action denoising latency, achieving denoising speedups of $6.62\times$ and $4.69\times$ on Fast-WAM and Faster-WAM, respectively, which translate into overall speedups of $4.21\times$ and $3.11\times$. For Joint-WAM, which jointly denoises video and actions, our method also accelerates the joint denoising process by $3.24\times$, yielding an overall speedup of $3.06\times$.

\begin{figure*}
	\centering
	\includegraphics[width=0.95\linewidth]{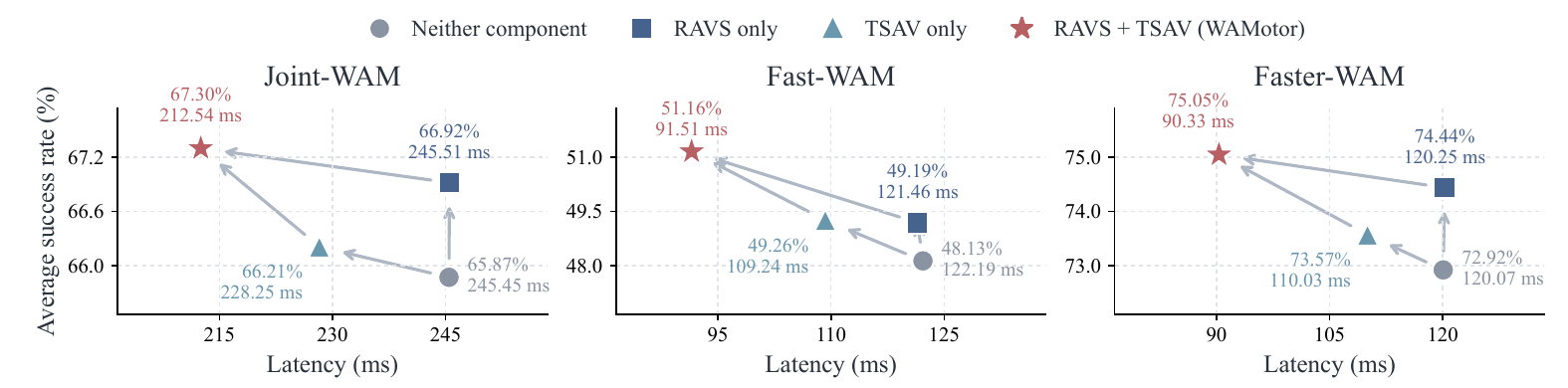}
	\caption{Component ablation results on LIBERO-Plus. Here, RAVS stands for Reliable Anchor Velocity Selection, and TSAV stands for Time-Step-Aware Verification.}
	\label{fig:component-ablation}
\end{figure*}

\begin{figure*}[t]
	\centering
	\begin{minipage}[t]{0.49\textwidth}
		\centering
		\includegraphics[width=\linewidth]{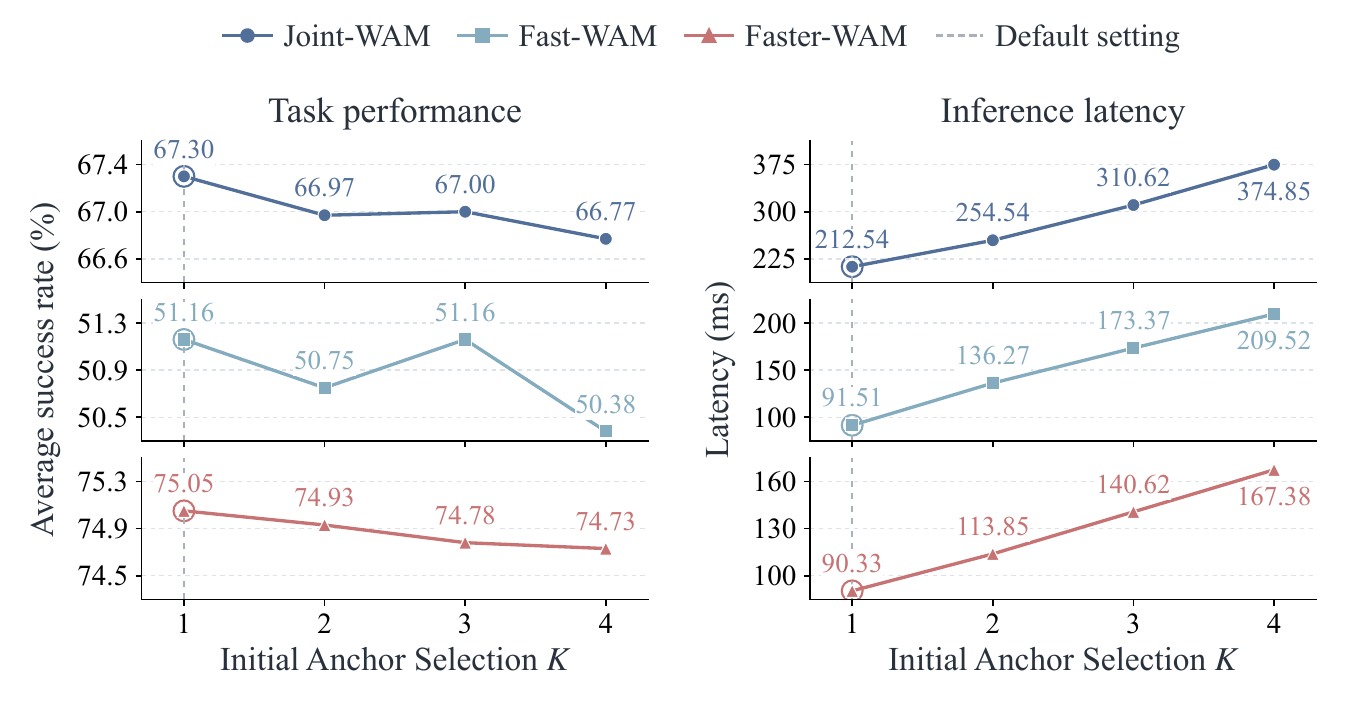}
		\caption{Sensitivity to initial anchor step $K$ on LIBERO-Plus.}
		\label{fig:sensitivity-k}
	\end{minipage}\hfill
	\begin{minipage}[t]{0.49\textwidth}
		\centering
		\includegraphics[width=\linewidth]{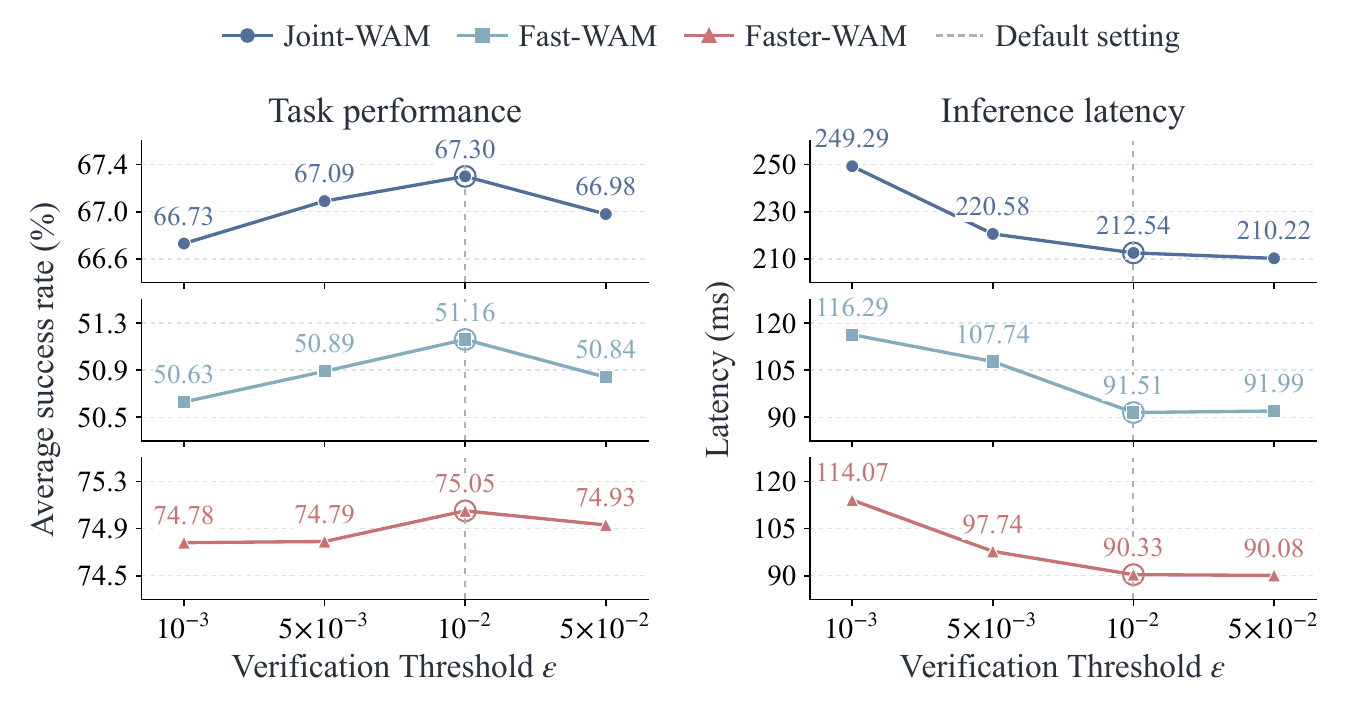}
		\caption{Sensitivity to verification threshold $\varepsilon$ on LIBERO-Plus.}
		\label{fig:sensitivity-epsilon}
	\end{minipage}
\end{figure*}

\subsection{Component Ablation}

To evaluate the contributions of Reliable Anchor Velocity Selection (RAVS) and Time-Step-Aware Verification (TSAV), we conduct an ablation study on LIBERO-Plus. As shown in \cref{fig:component-ablation}, RAVS alone improves average success rates by 1.05, 1.06, and 1.52 percentage points on the three WAMs, respectively, with nearly unchanged latency. TSAV alone improves success rates across all three models while reducing latency by 7.0\%--10.6\%. Combining both components achieves the highest average success rate and lowest latency on all three WAMs. Compared with disabling both components, the full method improves average success rates by 1.43, 3.03, and 2.13 percentage points, respectively, while reducing latency by 13.4\%--25.1\%. The complementary gains highlight the importance of jointly selecting reliable anchors and verifying their reuse according to the state perturbations induced by nonuniform integration time steps\footnote{Complete component ablation results in \cref{sec:detail-component-ablation}.}.

\subsection{Hyperparameter Sensitivity Analysis}

\paragraph{Initial Anchor Step $K$.} We vary the number of standard Euler updates before speculation over $K \in\{1,2,3,4\}$. As shown in \cref{fig:sensitivity-k}, $K=1$ achieves the highest average success rate and the lowest latency across all three models. Increasing $K$ beyond 1 increases action-chunk generation latency without improving average success rates. 
This supports our observations of velocity smoothness in WAM flow matching, showing that bypassing the first unreliable anchor improves both task performance and inference speed, whereas further bypassing subsequent reliable anchors diminishes these benefits\footnote{\label{fn:hyper}Complete hyperparameter sensitivity analysis results in \cref{sec:detail-hyperparameter-sensitivity}.}.

\paragraph{Verification Threshold $\varepsilon$.} We evaluate $\varepsilon$ from $10^{-3}$ to $5\times10^{-2}$. As shown in \cref{fig:sensitivity-epsilon}, reducing $\varepsilon$ from $10^{-2}$ to $10^{-3}$ increases latency across all three models without improving average success rates, consistent with the stricter local error budget in \cref{sec:threshold}. Conversely, increasing $\varepsilon$ to $5\times10^{-2}$ provides little additional latency benefit. Across the tested range, our method maintains competitive performance across a broad range of thresholds, with average success rates varying by at most 0.57 percentage points for each model, demonstrating robustness to the choice of $\varepsilon$\footref{fn:hyper}.

\section{Conclusion}

In this study, we uncover an overlooked dual nonuniformity in velocity-reuse speculative inference in WAMs: nonuniform velocity smoothness varies along denoising trajectories, and nonuniform integration time steps scale the resulting reuse errors, which accumulate into the final action distribution.
Base on this, we propose WAMotor, the first speculative inference method tailored to this dual nonuniformity in WAMs.
Guided by our observations and a theoretical link between velocity residuals, integration time steps, and final action distribution, WAMotor performs reliable anchor velocity selection and time-step-aware verification.
To demonstrate its plug-and-play benefits, we seamlessly integrate WAMotor into three representative WAMs and conduct evaluations on LIBERO, RoboTwin 2.0, and LIBERO-Plus. Compared with existing baselines, WAMotor consistently reduces latency while better preserving task performance. 
These results highlight the potential of velocity smoothness in WAMs to accelerate action generation, offering a path to efficient and reliable embodied control.
\section{LLM Usage Statement}

Large language models were used in a limited role during the preparation of this paper. We used GPT-6-Astra through Codex to assist with language editing to improve grammar, clarity, and readability, to assist with the implementation of the method code, and to provide an additional check of mathematical derivations that we had independently verified. We reviewed all AI-assisted content. We take full responsibility for the final content of this work, including any text, claims, or artifacts produced with the aid of large language model.
{
    \small
    \bibliographystyle{ieeenat_fullname}
    \bibliography{main}
}

\clearpage
\setcounter{page}{1}
\maketitlesupplementary
\appendix
\startcontents[supplementary]

In these Supplementary Materials, we provide additional details that could not be included in the main paper due to space limitations. 
They include the following:

\printcontents[supplementary]{}{1}{\setcounter{tocdepth}{2}}

\section{Theoretical Proofs}

\subsection{Proof of \cref{thm:traj}}
\label{sec:proof-thm-traj}

\begin{theorem*}
    Trajectory $\delta$-error tolerance can be given by a weighted sum of anchor velocity reuse residuals plus the propagated conditioning-error contribution, where larger residuals increase the bound with other terms held fixed, and earlier residuals carry no smaller propagation factors.
	
	\begin{equation}
        \begin{aligned}
            &\|a_N^s - a_N\|_{\mathcal{A}} \le \delta \\ := & \sum_{k \in \mathcal{I}} G_k\, h_k\, \|r_{M_k,k-M_k}\|_{\mathcal{A}} + B_{\mathrm{cond}},
        \end{aligned}
	\end{equation}
	
    where $\mathcal I$ denotes the set of update indices at which an earlier anchor velocity is reused in the transition from $k$ to $k+1$. For each $k\in\mathcal I$, $M_k<k$ denotes the index of the anchor whose velocity is reused.
\end{theorem*}

\begin{proof}
	We prove \cref{thm:traj} by induction on the per-step recursion for $\eta_k = \|a_k^s - a_k\|_{\mathcal{A}}$ with $\eta_0 = 0$.
	
	\textit{Step.} Consider the transition from $k$ to $k+1$. If $k\notin\mathcal I$, the speculative sampler performs an exact Euler update using the model velocity at its current state:
	\begin{equation}
        \begin{aligned}
    		a_{k+1}&=a_k+\Delta t_k\,v_\theta(a_k,t_k;c_k),\\
            a_{k+1}^s&=a_k^s+\Delta t_k\,v_\theta(a_k^s,t_k;c_k^s).
        \end{aligned}
    \end{equation}
	
	Subtracting the two updates and applying the triangle inequality and \cref{asm:lipschitz} gives
	\begin{equation}
		\begin{aligned}
			\eta_{k+1}
			&= \|a_{k+1}^s-a_{k+1}\|_{\mathcal A}\\
			&= \Big\|(a_k^s-a_k)
			+\Delta t_k\big[
			v_\theta(a_k^s,t_k;c_k^s)\\
			&\qquad\qquad
			-v_\theta(a_k,t_k;c_k)\big]\Big\|_{\mathcal A}\\
			&\le \|a_k^s-a_k\|_{\mathcal A}
			+|\Delta t_k|\,
			\big\|v_\theta(a_k^s,t_k;c_k^s)\\
			&\qquad\qquad
			-v_\theta(a_k,t_k;c_k)\big\|_{\mathcal A}\\
			&\le \eta_k
			+h_kL_k\|a_k^s-a_k\|_{\mathcal A} +h_kC_k\|c_k^s-c_k\|_{\mathcal A}\\
			&= \eta_k+h_kL_k\eta_k +h_kC_k\|c_k^s-c_k\|_{\mathcal A}\\
			&= (1+h_kL_k)\eta_k +h_kC_k\|c_k^s-c_k\|_{\mathcal A}.
		\end{aligned}
	\end{equation}
	
	If instead $k\in\mathcal I$, set $m=M_k$ and $j=k-M_k$. The current speculative state is $a_k^s=\hat a_{m,j}$, the speculative trajectory effectively reused velocity $u_m$ from step $m$ to $k$. The transition in the speculative trajectory can be written as
	
	\begin{equation}
		a_{k+1}^s=a_k^s+\Delta t_k\,u_m.
	\end{equation}
	
	By the definition of the verification residual,
    \begin{equation}
        \begin{aligned}
        r_{m,j}&=u_m-\hat v_{m,j},\\
		\hat v_{m,j}&=v_\theta(\hat a_{m,j},t_k;\hat c_{m,j})
		=v_\theta(a_k^s,t_k;c_k^s).
        \end{aligned}
	\end{equation}
	
	Hence
    \begin{equation}
        u_m=v_\theta(a_k^s,t_k;c_k^s)+r_{m,j},
    \end{equation}
    
    and subtracting the reference update yields
	\begin{equation}
		\begin{aligned}
			a_{k+1}^s-a_{k+1}
			&=(a_k^s-a_k)\\
			+&\Delta t_k\big[
			v_\theta(a_k^s,t_k;c_k^s)-v_\theta(a_k,t_k;c_k)\big]\\
			+&\Delta t_k\,r_{m,j}.
		\end{aligned}
	\end{equation}
	
	Applying the triangle inequality and the Lipschitz assumption,
	\begin{equation}
		\begin{aligned}
			\eta_{k+1}
			&= \|a_{k+1}^s-a_{k+1}\|_{\mathcal A}\\
			&= \Big\|(a_k^s-a_k)
			+\Delta t_k\big[
			v_\theta(a_k^s,t_k;c_k^s)\\
			&\qquad
			-v_\theta(a_k,t_k;c_k)\big]
			+\Delta t_k r_{m,j}\Big\|_{\mathcal A}\\
			&\le \|a_k^s-a_k\|_{\mathcal A}\\
			&\quad+|\Delta t_k|\,
			\|v_\theta(a_k^s,t_k;c_k^s)
			-v_\theta(a_k,t_k;c_k)\|_{\mathcal A}\\
			&\quad+|\Delta t_k|\,\|r_{m,j}\|_{\mathcal A}\\
			&\le \eta_k+h_kL_k\|a_k^s-a_k\|_{\mathcal A}
			+h_k\|r_{m,j}\|_{\mathcal A}\\
			&\quad +h_kC_k\|c_k^s-c_k\|_{\mathcal A}\\
			&= (1+h_kL_k)\eta_k
			+h_k\|r_{m,j}\|_{\mathcal A}\\
			&\quad +h_kC_k\|c_k^s-c_k\|_{\mathcal A}\\
			&= (1+h_kL_k)\eta_k
			+h_k\|r_{M_k,k-M_k}\|_{\mathcal A}\\
			&\quad +h_kC_k\|c_k^s-c_k\|_{\mathcal A},
		\end{aligned}
	\end{equation}
	
	To combine both cases without defining $M_k$ outside $\mathcal I$, introduce the proof-local residual
	\begin{equation}
		\rho_k:=
		\begin{cases}
			r_{M_k,k-M_k},& k\in\mathcal I,\\
			0,& k\notin\mathcal I.
		\end{cases}
	\end{equation}
	
	We therefore obtain the recursion
	\begin{equation}
		\label{eq:app-recursive}
        \begin{aligned}
            \eta_{k+1}&\le (1+h_kL_k)\eta_k+h_k\|\rho_k\|_{\mathcal A}\\
			&\quad+h_kC_k\|c_k^s-c_k\|_{\mathcal A}.
        \end{aligned}
	\end{equation}
	
	\textit{Unrolling.} Applying \cref{eq:app-recursive} repeatedly from $\eta_0=0$ gives
	\begin{equation}
		\begin{aligned}
			\eta_N
			&\le\sum_{k=0}^{N-1}
			\left(\prod_{\ell=k+1}^{N-1}(1+h_\ell L_\ell)\right)\\
            &\quad\times h_k\big(\|\rho_k\|_{\mathcal A}
			+C_k\|c_k^s-c_k\|_{\mathcal A}\big)\\
			&=\sum_{k=0}^{N-1}G_kh_k\|\rho_k\|_{\mathcal A}+B_{\mathrm{cond}}\\
			&=\sum_{k\in\mathcal I}
			G_kh_k\|r_{M_k,k-M_k}\|_{\mathcal A}+B_{\mathrm{cond}}.
		\end{aligned}
	\end{equation}
	
	The last equality follows because exact updates have $\rho_k=0$. This proves \cref{thm:traj}.
	
	Each residual magnitude enters this bound with a nonnegative coefficient $G_kh_k$. Moreover, for $k<N-1$, $G_k=(1+h_{k+1}L_{k+1})G_{k+1}\ge G_{k+1}$. Thus, earlier residuals carry no smaller propagation factors.
\end{proof}

\subsection{Proof of \cref{lem:local}}
\label{sec:proof-lem-local}

\begin{lemma*}
	The MSE between reusing the anchor velocity and performing an exact Euler update from the same speculative state scales quadratically with the integration time step size for a fixed velocity residual.
	
	\begin{equation}
		\begin{aligned}
			&\left\|
			(a_k^s+\Delta t_{k} u_{M_k})
			-(a_k^s+\Delta t_{k}\hat{v}_{M_k,k-M_k})
			\right\|_{\mathcal{A}}^2\\
			&\qquad=h_{k}^2\|r_{M_k,k-M_k}\|_{\mathcal{A}}^2.
		\end{aligned}
	\end{equation}
\end{lemma*}

\begin{proof}
	Subtracting the exact Euler update from the velocity-reuse update at $a_k^s$ gives
	\begin{equation}
		\begin{aligned}
			&(a_k^s+\Delta t_{k}u_{M_k})
			-(a_k^s+\Delta t_{k}\hat v_{M_k,k-M_k})\\
			&\qquad=\Delta t_{k}r_{M_k,k-M_k}.
		\end{aligned}
	\end{equation}
	
	Taking the norm and applying absolute homogeneity yields
	\begin{equation}
		\begin{aligned}
			&\|
			(a_k^s+\Delta t_{k}u_{M_k})
			-(a_k^s+\Delta t_{k}\hat v_{M_k,k-M_k})
			\|_{\mathcal A}\\
			&\qquad=|\Delta t_{k}|\|r_{M_k,k-M_k}\|_{\mathcal A}\\
			&\qquad=h_{k}\|r_{M_k,k-M_k}\|_{\mathcal A}.
		\end{aligned}
	\end{equation}
	
	Squaring both sides yields
	\begin{equation}
		\begin{aligned}
			&\|
			(a_k^s+\Delta t_{k}u_{M_k})
			-(a_k^s+\Delta t_{k}\hat v_{M_k,k-M_k})
			\|_{\mathcal A}^2\\
			&\qquad=h_{k}^2\|r_{M_k,k-M_k}\|_{\mathcal A}^2,
		\end{aligned}
	\end{equation}
	
	which proves \cref{eq:local-error}.
\end{proof}

\subsection{Proof of \cref{thm:time-bound}}
\label{sec:proof-thm-time-bound}

\begin{theorem*}
	Time-Step-Aware Verification with the median reference step $h^{\mathrm{mid}}$ and threshold $\varepsilon\ge0$ guarantees trajectory $\delta_{\mathrm{time}}$-error tolerance.
	
	\begin{equation}
		\|a_N^s - a_N\|_{\mathcal{A}} \;\le\;\delta_{\mathrm{time}}=\; h^{\mathrm{mid}}\sqrt{\varepsilon} \sum_{k\in\mathcal I}G_k + B_{\mathrm{cond}},
	\end{equation}
\end{theorem*}

\begin{proof}
	For each actual velocity-reuse update $k\in\mathcal I$, the time-step-aware acceptance criterion gives
	\begin{equation}
		\|r_{M_k,k-M_k}\|_{\mathcal A}^2
		\left(\frac{h_k}{h^{\mathrm{mid}}}\right)^2
		\le \varepsilon,
	\end{equation}
	
	and hence
	\begin{equation}
		h_k\|r_{M_k,k-M_k}\|_{\mathcal A}
		\le h^{\mathrm{mid}}\sqrt{\varepsilon}.
	\end{equation}

	By \cref{thm:traj}, each such perturbation contributes to the terminal discrepancy with propagation weight $G_k$. Since $G_k\ge0$, its weighted contribution satisfies
	\begin{equation}
		G_kh_k\|r_{M_k,k-M_k}\|_{\mathcal A}
		\le G_kh^{\mathrm{mid}}\sqrt{\varepsilon}.
	\end{equation}
	
	Summing over all actual velocity-reuse updates and applying \cref{thm:traj} yields
	\begin{equation}
		\|a_N^s-a_N\|_{\mathcal A}
		\le \delta_{\mathrm{time}}
		= h^{\mathrm{mid}}\sqrt{\varepsilon}
		\sum_{k\in\mathcal I}G_k+B_{\mathrm{cond}}.
	\end{equation}
    
	Therefore, the verification threshold $\varepsilon$ controls the local perturbation budget and, through the propagation weights, determines the stated trajectory error tolerance.
\end{proof}

\begin{table*}[t]
    \centering
    \small
    \setlength{\tabcolsep}{5.5pt}
    \renewcommand{\arraystretch}{1.05}
    \caption{Component ablation results on LIBERO-Plus across seven distribution shifts. Here, RAVS stands for Reliable Anchor Velocity Selection, and TSAV stands for Time-Step-Aware Verification.}
    \label{tab:component-ablation}
    \begin{tabular}{cccccccccccc}
    \toprule
    \multirow{2}{*}{Model}
    & \multicolumn{2}{c}{Component}
    & \multirow{2}{*}{Camera}
    & \multirow{2}{*}{Robot}
    & \multirow{2}{*}{Lang.}
    & \multirow{2}{*}{Light}
    & \multirow{2}{*}{Backg.}
    & \multirow{2}{*}{Noise}
    & \multirow{2}{*}{Layout}
    & \multirow{2}{*}{Avg. $\uparrow$}
    & \multirow{2}{*}{Lat. (ms) $\downarrow$} \\
    \cmidrule(lr){2-3}
    & \makecell{RAVS} & \makecell{TSAV} & & & & & & & & & \\
    \midrule

    \multirow{4}{*}{\rotatebox[origin=c]{90}{Joint-WAM}}
    & $\times$ & $\times$ & 35.74 & 61.92 & 90.37 & 90.37 & 54.29 & 49.37 & 79.00 & 65.87 & 245.45 \\
    & $\checkmark$ & $\times$ & 36.63 & 64.69 & 90.78 & 90.55 & 55.83 & 49.68 & 80.29 & \underline{66.92} & 245.51 \\
    & $\times$ & $\checkmark$ & 36.21 & 62.29 & 90.26 & 90.27 & 54.99 & 49.90 & 79.55 & 66.21 & \underline{228.25} \\
    & \cellcolor{cvprblue!20}$\checkmark$ & \cellcolor{cvprblue!20}$\checkmark$ & \cellcolor{cvprblue!20}36.64 & \cellcolor{cvprblue!20}64.97 & \cellcolor{cvprblue!20}90.77 & \cellcolor{cvprblue!20}90.32 & \cellcolor{cvprblue!20}56.38 & \cellcolor{cvprblue!20}50.01 & \cellcolor{cvprblue!20}82.02 & \cellcolor{cvprblue!20}\textbf{67.30} & \cellcolor{cvprblue!20}\textbf{212.54} \\

    \midrule

    \multirow{4}{*}{\rotatebox[origin=c]{90}{Fast-WAM}}
    & $\times$ & $\times$ & 15.15 & 43.47 & 67.42 & 80.33 & 43.98 & 28.42 & 58.12 & 48.13 & 122.19 \\
    & $\checkmark$ & $\times$ & 17.26 & 44.17 & 68.01 & 80.89 & 45.73 & 30.19 & 58.06 & 49.19 & 121.46 \\
    & $\times$ & $\checkmark$ & 17.47 & 44.33 & 67.92 & 80.89 & 45.99 & 30.02 & 58.20 & \underline{49.26} & \underline{109.24} \\
    & \cellcolor{cvprblue!20}$\checkmark$ & \cellcolor{cvprblue!20}$\checkmark$ & \cellcolor{cvprblue!20}18.84 & \cellcolor{cvprblue!20}47.35 & \cellcolor{cvprblue!20}69.05 & \cellcolor{cvprblue!20}82.30 & \cellcolor{cvprblue!20}48.65 & \cellcolor{cvprblue!20}31.71 & \cellcolor{cvprblue!20}60.20 & \cellcolor{cvprblue!20}\textbf{51.16} & \cellcolor{cvprblue!20}\textbf{91.51} \\

    \midrule

    \multirow{4}{*}{\rotatebox[origin=c]{90}{Faster-WAM}}
    & $\times$ & $\times$ & 52.98 & 68.86 & 95.08 & 92.82 & 57.51 & 61.28 & 81.88 & 72.92 & 120.07 \\
    & $\checkmark$ & $\times$ & 55.17 & 69.02 & 95.10 & 96.08 & 61.03 & 62.65 & 82.00 & \underline{74.44} & 120.25 \\
    & $\times$ & $\checkmark$ & 52.53 & 69.99 & 95.29 & 95.06 & 60.26 & 61.46 & 80.38 & 73.57 & \underline{110.03} \\
    & \cellcolor{cvprblue!20}$\checkmark$ & \cellcolor{cvprblue!20}$\checkmark$ & \cellcolor{cvprblue!20}55.48 & \cellcolor{cvprblue!20}69.31 & \cellcolor{cvprblue!20}96.30 & \cellcolor{cvprblue!20}95.46 & \cellcolor{cvprblue!20}60.70 & \cellcolor{cvprblue!20}64.99 & \cellcolor{cvprblue!20}83.13 & \cellcolor{cvprblue!20}\textbf{75.05} & \cellcolor{cvprblue!20}\textbf{90.33} \\

    \bottomrule
    \end{tabular}
\end{table*}

\begin{table*}[t]
    \fontsize{8pt}{10pt}\selectfont
    \centering
    \setlength{\tabcolsep}{1.2pt}
    \renewcommand{\arraystretch}{1.05}

    \begin{minipage}[t]{0.49\textwidth}
        \centering
        \caption{Sensitivity analysis of the initial anchor step $K$ on LIBERO-Plus across seven distribution shifts.}
        \label{tab:skip-step-sensitivity}
        \begin{tabular}{@{}cc*{9}{c}@{}}
        \toprule
        Model & $K$
        & Cam. & Rob. & Lang. & Light
        & Backg. & Noise & Layout & Avg. & Lat. \\
        \midrule

        \multirow{4}{*}{\rotatebox[origin=c]{90}{Joint-WAM}}
        & \cellcolor{cvprblue!20}1
        & \cellcolor{cvprblue!20}36.64 & \cellcolor{cvprblue!20}64.97 & \cellcolor{cvprblue!20}90.77 & \cellcolor{cvprblue!20}90.32
        & \cellcolor{cvprblue!20}56.38 & \cellcolor{cvprblue!20}50.01 & \cellcolor{cvprblue!20}82.02
        & \cellcolor{cvprblue!20}\textbf{67.30} & \cellcolor{cvprblue!20}\textbf{212.54} \\

        & 2
        & 36.59 & 64.12 & 91.01 & 90.40
        & 55.44 & 50.00 & 81.20
        & 66.97 & 254.54 \\

        & 3
        & 36.43 & 64.45 & 90.25 & 90.21
        & 55.97 & 50.02 & 81.69
        & 67.00 & 310.62 \\

        & 4
        & 36.40 & 63.91 & 90.81 & 90.12
        & 55.23 & 49.91 & 80.98
        & 66.77 & 374.85 \\

        \midrule

        \multirow{4}{*}{\rotatebox[origin=c]{90}{Fast-WAM}}
        & \cellcolor{cvprblue!20}1
        & \cellcolor{cvprblue!20}18.84 & \cellcolor{cvprblue!20}47.35 & \cellcolor{cvprblue!20}69.05 & \cellcolor{cvprblue!20}82.30
        & \cellcolor{cvprblue!20}48.65 & \cellcolor{cvprblue!20}31.71 & \cellcolor{cvprblue!20}60.20
        & \cellcolor{cvprblue!20}\textbf{51.16} & \cellcolor{cvprblue!20}\textbf{91.51} \\

        & 2
        & 18.26 & 47.01 & 69.10 & 81.91
        & 48.26 & 30.94 & 59.78
        & 50.75 & 136.27 \\

        & 3
        & 18.20 & 47.44 & 69.89 & 81.81
        & 48.70 & 31.79 & 60.31
        & \textbf{51.16} & 173.37 \\

        & 4
        & 17.58 & 46.88 & 68.38 & 82.07
        & 47.83 & 30.58 & 59.35
        & 50.38 & 209.52 \\

        \midrule

        \multirow{4}{*}{\rotatebox[origin=c]{90}{Faster-WAM}}
        & \cellcolor{cvprblue!20}1
        & \cellcolor{cvprblue!20}55.48 & \cellcolor{cvprblue!20}69.31 & \cellcolor{cvprblue!20}96.30 & \cellcolor{cvprblue!20}95.46
        & \cellcolor{cvprblue!20}60.70 & \cellcolor{cvprblue!20}64.99 & \cellcolor{cvprblue!20}83.13
        & \cellcolor{cvprblue!20}\textbf{75.05} & \cellcolor{cvprblue!20}\textbf{90.33} \\

        & 2
        & 54.46 & 70.34 & 95.71 & 95.29
        & 61.21 & 64.82 & 82.66
        & 74.93 & 113.85 \\

        & 3
        & 54.31 & 70.19 & 95.56 & 95.14
        & 61.06 & 64.67 & 82.51
        & 74.78 & 140.62 \\

        & 4
        & 54.37 & 70.09 & 95.50 & 95.10
        & 61.07 & 64.56 & 82.45
        & 74.73 & 167.38 \\

        \bottomrule
        \end{tabular}
    \end{minipage}
    \hfill
    \begin{minipage}[t]{0.49\textwidth}
        \centering
        \caption{Sensitivity analysis of verification threshold $\varepsilon$ on LIBERO-Plus across seven distribution shifts.}
        \label{tab:threshold-sensitivity}
        \begin{tabular}{@{}cc*{9}{c}@{}}
        \toprule
        Model & $\varepsilon$
        & Cam. & Rob. & Lang. & Light
        & Backg. & Noise & Layout & Avg. & Lat. \\
        \midrule

        \multirow{4}{*}{\rotatebox[origin=c]{90}{Joint-WAM}}
        & $5e-2$
        & 36.41 & 64.53 & 90.42 & 90.03
        & 56.39 & 49.39 & 81.70
        & 66.98 & \textbf{210.22} \\

        & \cellcolor{cvprblue!20}$1e-2$
        & \cellcolor{cvprblue!20}36.64 & \cellcolor{cvprblue!20}64.97 & \cellcolor{cvprblue!20}90.77 & \cellcolor{cvprblue!20}90.32
        & \cellcolor{cvprblue!20}56.38 & \cellcolor{cvprblue!20}50.01 & \cellcolor{cvprblue!20}82.02
        & \cellcolor{cvprblue!20}\textbf{67.30} & \cellcolor{cvprblue!20}212.54 \\

        & $5e-3$
        & 36.62 & 64.64 & 90.43 & 90.15
        & 56.51 & 49.46 & 81.83
        & 67.09 & 220.58 \\

        & $1e-3$
        & 36.42 & 64.14 & 90.43 & 90.10
        & 55.47 & 49.27 & 81.29
        & 66.73 & 249.29 \\

        \midrule

        \multirow{4}{*}{\rotatebox[origin=c]{90}{Fast-WAM}}
        & $5e-2$
        & 18.08 & 46.94 & 69.12 & 81.98
        & 48.39 & 31.42 & 59.93
        & 50.84 & 91.99 \\

        & \cellcolor{cvprblue!20}$1e-2$
        & \cellcolor{cvprblue!20}18.84 & \cellcolor{cvprblue!20}47.35 & \cellcolor{cvprblue!20}69.05 & \cellcolor{cvprblue!20}82.30
        & \cellcolor{cvprblue!20}48.65 & \cellcolor{cvprblue!20}31.71 & \cellcolor{cvprblue!20}60.20
        & \cellcolor{cvprblue!20}\textbf{51.16} & \cellcolor{cvprblue!20}\textbf{91.51} \\

        & $5e-3$
        & 18.27 & 46.90 & 69.13 & 82.17
        & 48.37 & 31.56 & 59.81
        & 50.89 & 107.74 \\

        & $1e-3$
        & 18.06 & 46.50 & 68.81 & 82.00
        & 48.24 & 31.22 & 59.56
        & 50.63 & 116.29 \\

        \midrule

        \multirow{4}{*}{\rotatebox[origin=c]{90}{Faster-WAM}}
        & $5e-2$
        & 54.31 & 69.31 & 95.92 & 96.01
        & 61.01 & 64.84 & 83.12
        & 74.93 & \textbf{90.08} \\

        & \cellcolor{cvprblue!20}$1e-2$
        & \cellcolor{cvprblue!20}55.48 & \cellcolor{cvprblue!20}69.31 & \cellcolor{cvprblue!20}96.30 & \cellcolor{cvprblue!20}95.46
        & \cellcolor{cvprblue!20}60.70 & \cellcolor{cvprblue!20}64.99 & \cellcolor{cvprblue!20}83.13
        & \cellcolor{cvprblue!20}\textbf{75.05} & \cellcolor{cvprblue!20}90.33 \\

        & $5e-3$
        & 54.36 & 69.35 & 95.71 & 95.70
        & 60.67 & 64.75 & 82.96
        & 74.79 & 97.74 \\

        & $1e-3$
        & 54.29 & 69.26 & 95.74 & 95.61
        & 60.78 & 64.83 & 82.92
        & 74.78 & 114.07 \\

        \bottomrule
        \end{tabular}
    \end{minipage}
\end{table*}

\section{Supporting Information for Main-Text}
\label{sec:supp-info}

\subsection{Additional Related Work}
\label{sec:addition-related-work}

\paragraph{Caching-based Acceleration for Diffusion Models.}

Caching-based acceleration methods reduce the computational cost of iterative denoising by reusing intermediate network activations, such as hidden states and attention key--value pairs, across adjacent steps. Following early successes in visual diffusion models~\cite{deepcache-ma,teacache-liu,toca-zou,speca-liu}, recent works exemplified by VLA-Cache~\cite{vla-cache-xu} have adapted feature and token caching to embodied policies~\cite{bac-ji,wam-cache-ding,efficientvla-yang,c3cache-zhao,actioncache-oi,fast-vla-gong,ttf-vla-liu}. Despite their efficiency, these methods predominantly rely on feature-level similarity metrics or predefined schedules to govern reuse, lacking a verification mechanism to bound the resulting deviation in the final action distribution. In the main paper, we benchmark WAMotor against representative caching methods to demonstrate the advantage of our error-bounded speculative framework. Meanwhile, because internal feature caching and trajectory-level velocity speculation operate at complementary levels of granularity, they are orthogonal in design. We empirically validate this synergy in \cref{sec:complementarity}, showing that stacking WAMotor onto existing caching methods yields additional speedups while improving task success rates.

\subsection{Detailed Information of Benchmark}
\label{sec:detail-bench}

\paragraph{LIBERO.} LIBERO~\cite{libero-liu} is a standard benchmark for language-conditioned manipulation, comprising four suites that cover spatial relations, object-centric skills, goal-conditioned tasks, and long-horizon behaviors. LIBERO-Spatial distinguishes identical objects by their spatial relations, LIBERO-Object requires picking and placing a distinct object in each task, LIBERO-Goal shares a fixed scene but varies the task goal, and LIBERO-Long contains long-horizon tasks with diverse object interactions. Each suite contains 10 tasks, and each task is evaluated over 50 trials.

\paragraph{RoboTwin 2.0.} RoboTwin 2.0~\cite{robotwin2.0-chen} is a large-scale bimanual manipulation benchmark with 50 tasks requiring coordinated dual-arm control under diverse scene conditions. It is built on an object library of 731 instances across 147 categories and applies domain randomization over clutter, lighting, backgrounds, tabletop height, and language instructions. Each task is evaluated over 100 trials in both clean and randomized settings.

\paragraph{LIBERO-Plus.} LIBERO-Plus~\cite{libero-plus-fei} extends the original LIBERO tasks to evaluate robustness under conditions not observed during standard training. It introduces seven types of variation involving camera viewpoints, robot initial states, language instructions, lighting, backgrounds, sensor noise, and object layouts, yielding 10,030 perturbed tasks in total. Each task is evaluated over 3 trials.

\subsection{Detailed Results of Component Ablation}
\label{sec:detail-component-ablation}

The main paper reports average success rates and inference latency on LIBERO-Plus under different combinations of Reliable Anchor Velocity Selection (RAVS) and Time-Step-Aware Verification (TSAV). Here, we supplement these results with a detailed breakdown of task success rates across seven distribution shifts for all three WAMs under each setting. Compared with disabling both components, combining RAVS and TSAV improves success rates across most of the seven distribution shifts for each WAM. Notable gains include robot initial-state and object-layout perturbations on Joint-WAM (+3.05 pp and +3.02 pp), lighting and background perturbations on Fast-WAM (+1.97 pp and +4.67 pp), and camera and sensor-noise perturbations on Faster-WAM (+2.50 pp and +3.71 pp). The complete results are presented in \cref{tab:component-ablation}.

\subsection{Detailed Results of Hyperparameter Sensitivity Analysis}
\label{sec:detail-hyperparameter-sensitivity}

The main paper reports average success rates and inference latency on LIBERO-Plus under different settings of initial anchor step $K$ and verification threshold $\varepsilon$. Here, we supplement these results with a detailed breakdown of task success rates across seven distribution shifts for all three WAMs under each setting. Across most of the individual distribution shifts, increasing $K$ yields no improvement. For example, increasing $K$ from 1 to 4 reduces success rates under camera and object-layout perturbations for all three WAMs. Performance is also stable across the tested verification thresholds. 
Furthermore, success rates under each distribution shift remain relatively stable across the tested values of $\varepsilon$.
These results further support the robustness of the default settings, $K=1$ and $\varepsilon=10^{-2}$. The complete results are presented in \cref{tab:skip-step-sensitivity} and \cref{tab:threshold-sensitivity}.

\begin{algorithm}
    \caption{WAMotor (joint video--action prediction)}
    \label{algorithm-joint}
    \begin{algorithmic}[1]
        \Require Gaussian noise $a_0^s$, Initial future state $c_0^s$, Diffusion step $N$
        \Ensure Action $a_N^s$
        \State $\left(v_0^c, v_0^a\right) \gets v_\theta \left(a_0^s,t_0;c_0^s\right)$ \Comment{Stage I}
        \State $c_1^s \gets c_0^s + \Delta t_0^c\,v_0^c$, \quad $a_1^s \gets a_0^s + \Delta t_0\,v_0^a$
        \State $m \gets 1$, \quad $\left(u_m^c, u_m\right) \gets v_\theta(a_m^s,t_m;c_m^s)$
        \While{$m < N - 1$}
            \State $\hat c_{m,0}=c_m^s$, $\hat a_{m,0}=a_m^s$, $\hat v_{m,0}^c=u_m^c$, $\hat v_{m,0}=u_m$
            \State \textbf{Draft (Future):} $\hat{c}_{m,j} = c_m^s + u_m^c\sum\limits_{i=0}^{j-1}\Delta t_{m+i}^c$,\, $1 \le j \le q_m$
            \State \textbf{Draft (Action):} $\hat{a}_{m,j} = a_m^s + u_m\sum\limits_{i=0}^{j-1}\Delta t_{m+i}$,\, $1 \le j \le q_m$
            \State \textbf{Verify:} $\left(\hat{v}_{m,j}^c, \hat{v}_{m,j}\right) \gets v_\theta\left(\hat{a}_{m,j},t_{m+j};\hat c_{m,j}\right)$, $e_{m+j} \gets \|u_m-\hat v_{m,j}\|_{\mathcal A}^2\cdot\left(\frac{h_{m+j}}{h^{\mathrm{mid}}}\right)^2$, $1 \le j \le q_m$ \Comment{Stage II}
            \State Find first index $i \in [1,q_m]$ such that $e_{m+i} > \varepsilon$
            \If{$e_{m+j} \le \varepsilon$, $1 \le j \le q_m$}
                \State $c_{m+j}^s \gets \hat{c}_{m,j}$, \quad $a_{m+j}^s \gets \hat{a}_{m,j}$, $1 \le j \le q_m$
                \State $u_{m+q_m}^c \gets \hat{v}_{m,q_m}^c$, \quad $u_{m+q_m} \gets \hat{v}_{m,q_m}$, \quad $m \gets m + q_m$
            \Else
                \State \textbf{Correction:} $c_{m+j}^s \gets \hat{c}_{m,j}$, \quad $a_{m+j}^s \gets \hat{a}_{m,j}$, $1 \le j < i$
                \State $c_{m+i}^s\gets \hat c_{m,i-1}+\Delta t_{m+i-1}^c\hat v_{m,i-1}^c$, \quad $a_{m+i}^s\gets \hat a_{m,i-1}+\Delta t_{m+i-1}\hat v_{m,i-1}$
                \State $\left(u_{m+i}^c, u_{m+i}\right)\gets v_\theta(a_{m+i}^s,t_{m+i};c_{m+i}^s)$, \quad $m \gets m + i$
            \EndIf
        \EndWhile
        \State $a_N^s \gets a_{N-1}^s+\Delta t_{N-1}\,u_{N-1}$
        \State \Return $a_N^s$
    \end{algorithmic}
\end{algorithm}

\subsection{Pseudocode for the Integration of WAMotor in Joint-WAMs}
\label{sec:pseudocode-joint}

\begin{figure*}
	\centering
	\includegraphics[width=0.9\linewidth]{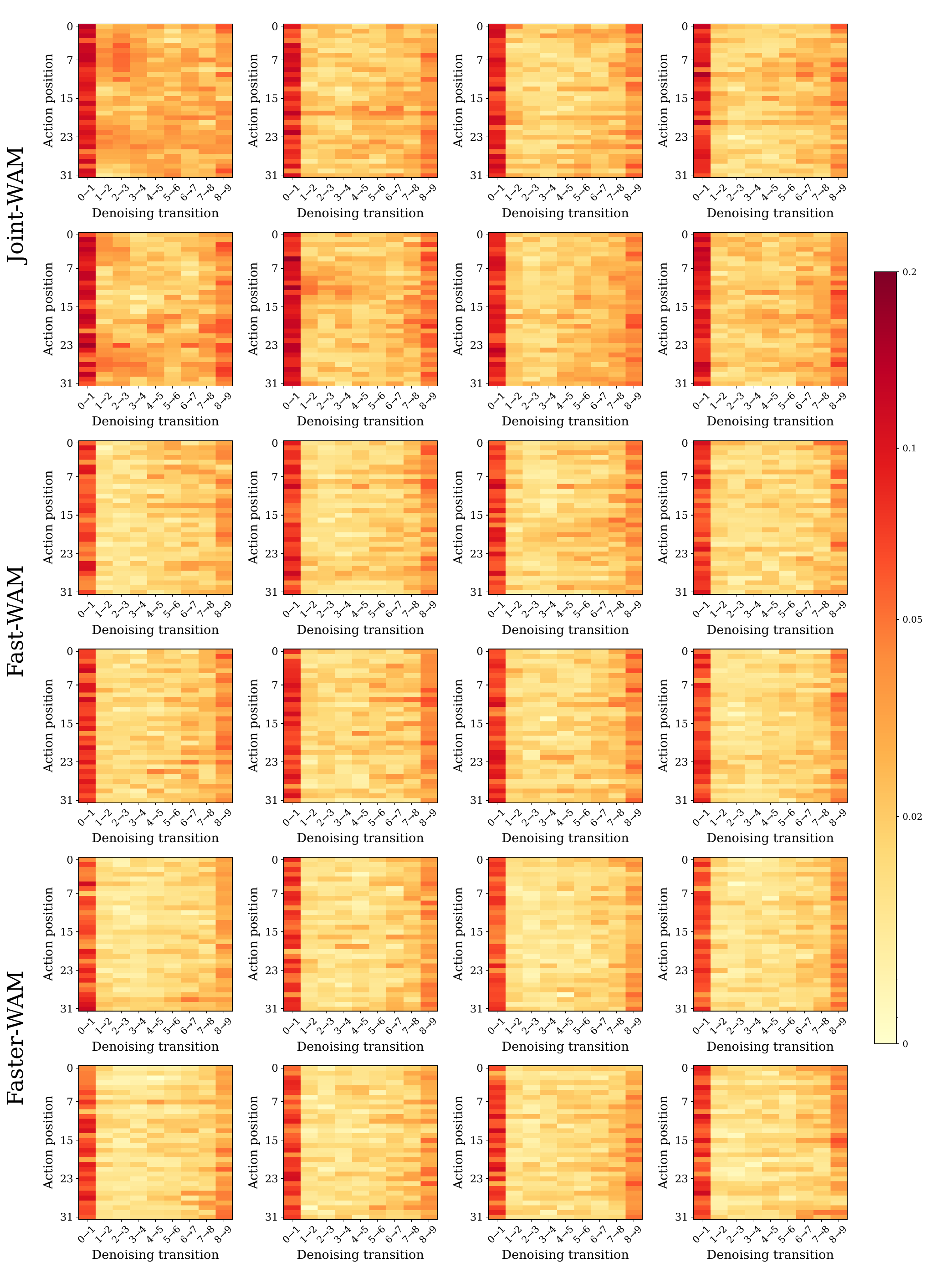}
	\caption{Visualization of velocity variation across denoising steps.}
	\label{fig:caseheatmaps}
\end{figure*}

In the main paper, \cref{algorithm} presents the general procedure of WAMotor for action generation under conditioning context $c_k^s$ (\eg Fast-WAM and Faster-WAM). Due to space constraints in the main text, we provide the detailed pseudocode in \cref{algorithm-joint}, which describes how to integrate WAMotor into WAMs with explicit future video prediction (\eg Joint-WAM). Unlike action-only or cached-conditioning variants where the conditioning context remains fixed across denoising steps, Joint-WAM jointly denoises future video latents and action states at each step. 

Accordingly, WAMotor extends velocity reuse to both modalities. Following Reliable Anchor Velocity Selection, it first conducts one standard Euler update on both future states and actions to obtain $c_1^s$ and $a_1^s$, using their velocities $\left(u_1^c, u_1\right)$ as the initial anchors. At each subsequent anchor step $m$, it simultaneously drafts candidate future states $\hat{c}_{m,j}$ and candidate actions $\hat{a}_{m,j}$ using their respective anchor velocities $u_m^c$ and $u_m$. During parallel verification, the model jointly evaluates the velocities $\left(\hat{v}_{m,j}^c, \hat{v}_{m,j}\right)$ over the drafted pairs, while acceptance is governed by the time-step-aware verification on the action velocity residuals. Once the longest accepted prefix is determined, both the speculative future states $c_{m+j}^s$ and actions $a_{m+j}^s$ are synchronously committed and corrected, ensuring consistent video--action evolution throughout the speculative trajectory.

\section{Extended Analysis Experiments}
\label{sec:ex-analysis}

\subsection{Detailed Visualization of Smoothness in WAM’s Flow Matching}
\label{sec:heatmap-visualization}

\Cref{fig:caseheatmaps} presents additional visualizations of velocity variation across consecutive denoising steps for Joint-WAM, Fast-WAM, and Faster-WAM on selected LIBERO-Plus tasks, supplementing \cref{fig:motivation-left}. Each heatmap shows the norm of the difference between consecutive velocity predictions at each action position within a chunk. Across all three models, the transition between the first two velocity predictions exhibits consistently pronounced differences, whereas most intermediate transitions show smaller variations. This pattern supports our reliable anchor velocity selection strategy, which performs one standard Euler update before using the updated velocity as the initial speculation anchor.

\subsection{Quantitative Analysis of Nonuniform Velocity Smoothness}
\label{sec:app-velocity-stats}

To quantitatively corroborate our observation in \cref{sec:motivation-1} that velocity smoothness varies across denoising steps, we evaluate the Endpoint MSE Share across all 40 LIBERO tasks for Joint-WAM, Fast-WAM, and Faster-WAM. Given the adjacent velocity MSE $m_k = \mathrm{MSE}(v_{k},v_{k+1})$ at each transition ($k=1,\dots,N-1$), the Endpoint MSE Share is defined as

\begin{equation}
    S_{\mathrm{end}} = S_{\mathrm{first}} + S_{\mathrm{last}} = \frac{m_1 + m_{N-1}}{\sum_{k=1}^{N-1} m_k},
\end{equation}

where $S_{\mathrm{first}} = \frac{m_1}{\sum_{k=1}^{N-1} m_k}$ and $S_{\mathrm{last}} = \frac{m_{N-1}}{\sum_{k=1}^{N-1} m_k}$ denote the proportions of total velocity MSE at the first and last transitions, respectively.

\begin{table}
    \centering
    \small
    \setlength{\tabcolsep}{3pt}
    \caption{Quantitative statistics of Endpoint MSE Share on LIBERO. For comparison, Uniform Ref. denotes the uniform case where all transitions contribute equally, so each transition accounts for $1/(N-1) \approx 11.11\%$ of the total MSE, yielding $S_{\mathrm{end}} \approx 22.22\%$.}
    \label{tab:velocity-endpoint-stats}
    \begin{tabular}{lccc}
        \toprule
        Model & $S_{\mathrm{first}}$ (\%) & $S_{\mathrm{last}}$ (\%) & $S_{\mathrm{end}}$ (\%) \\
        \midrule
        Uniform Ref. & 11.11 & 11.11 & 22.22 \\
        \midrule
        Joint-WAM & 41.21 $\pm$ 6.82 & 25.22 $\pm$ 5.52 & 66.43 $\pm$ 6.93 \\
        Fast-WAM & 43.59 $\pm$ 5.67 & 36.29 $\pm$ 6.63 & 79.88 $\pm$ 2.49 \\
        Faster-WAM & 48.07 $\pm$ 7.79 & 30.91 $\pm$ 6.53 & 78.98 $\pm$ 3.76 \\
        \bottomrule
    \end{tabular}
\end{table}

\begin{table*}
	\centering
	\small
	\setlength{\tabcolsep}{4pt}
	\caption{Sensitivity to reference step selection across integration time step schedules on Fast-WAM, evaluated on LIBERO-Plus. Min, Max, Mean, Q25, Q75, and Median denote the minimum, maximum, mean, 25th percentile, 75th percentile, and median of the integration step sizes, respectively.}
	\label{tab:sensitivity-reference-step}
	\begin{tabular}{llccccccccc}
		\toprule
		Schedules & Reference
		& Camera & Robot & Lang. & Light & Backg. & Noise & Layout
		& Avg. $\uparrow$ & Lat. (ms) $\downarrow$ \\
		\midrule
		\multirow{2}{*}{Uniform} & Vanilla & 14.66 & 46.31 & 68.02 & 81.10 & 47.17 & 33.32 & 59.81 & 50.06 & 395.89 \\
		  & \cellcolor{cvprblue!20}Median & \cellcolor{cvprblue!20}16.67 & \cellcolor{cvprblue!20}48.32 & \cellcolor{cvprblue!20}66.12 & \cellcolor{cvprblue!20}81.38 & \cellcolor{cvprblue!20}47.32 & \cellcolor{cvprblue!20}34.08 & \cellcolor{cvprblue!20}56.61 & \cellcolor{cvprblue!20}50.07 & \cellcolor{cvprblue!20}93.77 \\
		 
		\midrule
		 
		\multirow{7}{*}{Shift-5} & Vanilla & 16.53 & 46.11 & 68.03 & 82.72 & 47.94 & 30.74 & 59.09 & 50.17 & 391.43 \\
		  & Min & 13.18 & 46.04 & 65.01 & 80.18 & 48.17 & 31.85 & 59.91 & 49.19 & 100.31 \\
		  & Max & 15.67 & 49.37 & 66.40 & 78.65 & 44.47 & 34.32 & 60.81 & 49.96 & 89.75 \\
		  & Mean & 14.42 & 47.32 & 65.11 & 80.38 & 51.87 & 35.55 & 60.73 & 50.77 & 93.15 \\
		  & Q25 & 13.17 & 48.61 & 69.00 & 78.67 & 55.58 & 29.38 & 60.84 & 50.75 & 95.07 \\
		  & Q75 & 12.67 & 47.46 & 65.78 & 77.20 & 46.32 & 34.32 & 62.11 & 49.41 & 89.30 \\
		  & \cellcolor{cvprblue!20}Median &	\cellcolor{cvprblue!20}18.84 & \cellcolor{cvprblue!20}47.35 & \cellcolor{cvprblue!20}69.05 & \cellcolor{cvprblue!20}82.30 & \cellcolor{cvprblue!20}48.65 & \cellcolor{cvprblue!20}31.71 & \cellcolor{cvprblue!20}60.20 & \cellcolor{cvprblue!20}51.16 & \cellcolor{cvprblue!20}91.51 \\
		 
		  \midrule
		 
        \multirow{7}{*}{Shift-0.2} & Vanilla & 14.42 & 42.20 & 61.21 & 82.10 & 50.02 & 29.38 & 58.26 & 48.23 & 391.41 \\
        & Min & 13.10 & 42.25 & 61.22 & 78.65 & 48.17 & 24.44 & 55.61 & 46.21 & 99.53 \\
        & Max & 11.92 & 47.36 & 60.82 & 76.93 & 48.01 & 28.15 & 59.55 & 47.53 & 88.79 \\
        & Mean & 13.17 & 43.48 & 62.51 & 78.63 & 48.17 & 25.67 & 55.62 & 46.75 & 90.45 \\
        & Q25 & 13.47 & 40.91 & 59.91 & 82.08 & 51.87 & 29.33 & 58.21 & 47.97 & 95.07 \\
        & Q75 & 10.67 & 42.22 & 57.31 & 76.93 & 55.58 & 31.83 & 56.97 & 47.36 & 91.03 \\
        & \cellcolor{cvprblue!20}Median & \cellcolor{cvprblue!20}16.67 & \cellcolor{cvprblue!20}47.89 & \cellcolor{cvprblue!20}65.15 & \cellcolor{cvprblue!20}78.61 & \cellcolor{cvprblue!20}50.02 & \cellcolor{cvprblue!20}31.85 & \cellcolor{cvprblue!20}58.27 & \cellcolor{cvprblue!20}49.78 & \cellcolor{cvprblue!20}91.68 \\
		 
        \midrule
        
        \multirow{7}{*}{Cosine} & Vanilla & 13.17 & 44.76 & 66.40 & 80.38 & 50.96 & 28.15 & 56.91 & 48.68 & 399.27 \\
        & Min & 14.35 & 42.29 & 65.11 & 78.54 & 55.58 & 26.91 & 58.22 & 48.71 & 99.65 \\
        & Max & 15.67 & 47.33 & 61.26 & 78.32 & 53.73 & 26.91 & 54.33 & 48.22 & 89.76 \\
        & Mean & 14.27 & 44.76 & 65.30 & 75.20 & 51.87 & 24.44 & 58.20 & 47.72 & 92.74 \\
        & Q25 & 14.42 & 46.04 & 65.12 & 78.77 & 53.73 & 28.15 & 56.95 & 49.03 & 95.01 \\
        & Q75 & 15.67 & 51.17 & 58.61 & 76.93 & 48.11 & 30.63 & 60.82 & 48.85 & 89.39 \\
        & \cellcolor{cvprblue!20}Median & \cellcolor{cvprblue!20}18.17 & \cellcolor{cvprblue!20}48.61 & \cellcolor{cvprblue!20}63.81 & \cellcolor{cvprblue!20}75.20 & \cellcolor{cvprblue!20}53.73 & \cellcolor{cvprblue!20}30.61 & \cellcolor{cvprblue!20}58.56 & \cellcolor{cvprblue!20}49.81 & \cellcolor{cvprblue!20}92.66 \\
		\bottomrule
	\end{tabular}
\end{table*}

\begin{figure*}
	\centering
	\includegraphics[width=1\linewidth]{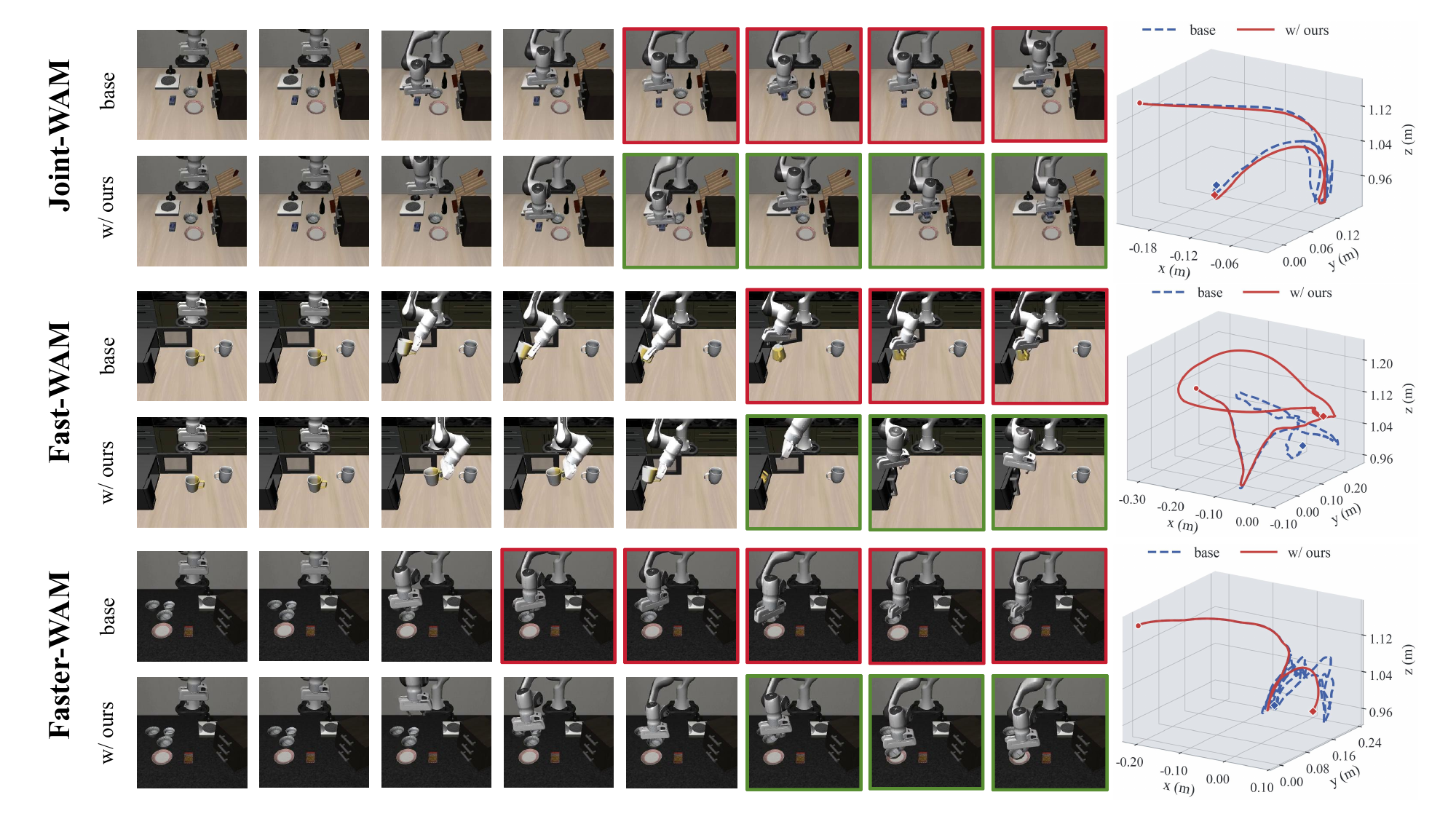}
	\caption{Case Study of Action Trajectories with and without Our Method. The red boxes indicate the trajectories in which the corresponding base model produced errors, while the green boxes indicate the trajectories that became correct after incorporating our method. The 3D plot on the right provides a comparison of their complete 3D trajectories.}
	\label{fig:case-study}
\end{figure*}

As shown in \cref{tab:velocity-endpoint-stats}, the two boundary transitions account for $66.43\%$, $79.88\%$, and $78.98\%$ of the total MSE on Joint-WAM, Fast-WAM, and Faster-WAM, respectively—over $2.9\times$ the $22.22\%$ uniform baseline. Notably, the first transition alone contributes $41.21\%$--$48.07\%$ of the total MSE, leaving the seven intermediate transitions with only $20.12\%$--$33.57\%$ in total. This pronounced boundary concentration confirms that velocity variations occur primarily at the beginning and end of denoising while intermediate transitions remain highly smooth, directly validating our Reliable Anchor Velocity Selection strategy.

\subsection{Sensitivity to Reference Step across Integration Time Step Schedules}
\label{sec:sensitivity-ref}


\Cref{tab:sensitivity-reference-step} compares reference step choices across four integration time step schedules on Fast-WAM: Uniform uses constant step sizes. Shift-5 and Shift-0.2 apply rational time shifting with factors of 5 and 0.2, producing increasing and decreasing step sizes, respectively. Cosine applies cosine time discretization, yielding smaller steps at both ends and larger steps in the middle~\cite{iddpm-quinn}. Shift-5 is the default schedule for Joint-WAM, Fast-WAM, and Faster-WAM. Under the uniform schedule, all reference choices coincide, so only the median is reported. 

Across all four schedules, median calibration achieves the highest average success rate while maintaining near-minimal latency. Larger reference values, such as the maximum or upper quartile, sometimes yield slightly lower latency but reduce average success rates, whereas the minimum increases latency without improving performance. By using a central reference step, median calibration avoids overly tightening or relaxing verification and is less sensitive to extremely large or small step sizes than mean- or extrema-based calibration. These results support the median as a reliable default that balances task performance and inference efficiency across different integration schedules.

\begin{table}
	\centering
	\small
	\setlength{\tabcolsep}{10pt}
	\caption{Action entropy measured on LIBERO-Plus across different models and inference settings. Each model uses 50 uniformly sampled cases, paired across settings. Values are mean $\pm$ sample standard deviation (in bits).}
	\label{tab:speculative-entropy}
	\begin{tabular}{lcc}
		\toprule
		\multirow{2}{*}{Model} & \multicolumn{2}{c}{Action Entropy (bits)} \\
		\cmidrule(lr){2-3}
		& Vanilla & w/ ours \\
		\midrule
		Joint-WAM  & $2.31 \pm 0.19$ & $2.01 \pm 0.22$ \\
		Fast-WAM   & $2.15 \pm 0.17$ & $1.80 \pm 0.17$ \\
		Faster-WAM & $2.09 \pm 0.36$ & $1.87 \pm 0.35$ \\
		\bottomrule
	\end{tabular}
\end{table}

\begin{table*}[t]
	\centering
	\small
	\setlength{\tabcolsep}{4pt}
	\renewcommand{\arraystretch}{1.05}
	\caption{Complementarity of WAMotor with diffusion caching methods on LIBERO-Plus across seven distribution shifts. Each row applies WAMotor on top of the corresponding caching baseline. $\Delta$ denotes the change relative to the same caching baseline without WAMotor (\cref{tab:libero_plus}), in percentage points for Avg. and in ms for Lat.}
	\label{tab:caching-complementarity}
	\begin{tabular}{clccccccccccc}
		\toprule
		\multirow{2}{*}{Model} & \multirow{2}{*}{Baseline}
		& \multirow{2}{*}{Camera} & \multirow{2}{*}{Robot} & \multirow{2}{*}{Lang.} & \multirow{2}{*}{Light} & \multirow{2}{*}{Backg.} & \multirow{2}{*}{Noise} & \multirow{2}{*}{Layout}
		& \multicolumn{2}{c}{Avg. $\uparrow$} & \multicolumn{2}{c}{Lat. (ms) $\downarrow$} \\
		\cmidrule(lr){10-11} \cmidrule(lr){12-13}
		& & & & & & & & & Value & $\Delta$ & Value & $\Delta$ \\
		\midrule
		
		\multirow{4}{*}{\rotatebox[origin=c]{90}{Joint-WAM}}
		& VLA-Cache
		& 36.52 & 61.10 & 89.73 & 90.22 & 52.76 & 48.11 & 78.54 & 65.28 & \dpos{2.65} & 202.54 & \dneg{230.03} \\
		& TeaCache
		& 36.76 & 64.67 & 90.76 & 90.22 & 56.28 & 49.94 & 82.08 & 67.24 & \dpos{1.91} & 180.74 & \dneg{104.60} \\
		& BAC
		& 36.59 & 63.58 & 90.72 & 90.33 & 55.62 & 49.70 & 81.13 & 66.81 & \dpos{0.83} & 194.61 & \dneg{126.97} \\
		& SpeCa
		& 36.79 & 60.21 & 88.96 & 88.76 & 53.77 & 48.25 & 78.30 & 65.01 & \dpos{3.28} & 175.58 & \dneg{83.89} \\
		
		\midrule
		
		\multirow{4}{*}{\rotatebox[origin=c]{90}{Fast-WAM}}
		& VLA-Cache
		& 16.98 & 47.36 & 59.41 & 81.86 & 47.89 & 31.54 & 59.42 & 49.21 & \dpos{2.06} & 85.92 & \dneg{120.23} \\
		& TeaCache
		& 17.09 & 46.21 & 66.27 & 78.97 & 46.61 & 31.71 & 59.33 & 49.46 & \dpos{2.15} & 70.99 & \dneg{62.39} \\
		& BAC
		& 18.38 & 46.90 & 68.54 & 81.85 & 48.84 & 30.86 & 60.37 & 50.82 & \dpos{1.00} & 80.93 & \dneg{79.78} \\
		& SpeCa
		& 18.01 & 44.99 & 68.70 & 80.99 & 47.73 & 29.59 & 59.53 & 49.93 & \dpos{0.42} & 65.05 & \dneg{51.95} \\
		
		\midrule
		
		\multirow{4}{*}{\rotatebox[origin=c]{90}{Faster-WAM}}
		& VLA-Cache
		& 52.96 & 68.99 & 94.17 & 92.88 & 58.81 & 63.15 & 82.49 & 73.35 & \dpos{3.03} & 85.44 & \dneg{112.84} \\
		& TeaCache
		& 54.18 & 70.21 & 95.41 & 94.68 & 60.32 & 63.21 & 83.03 & 74.43 & \dpos{1.02} & 69.58 & \dneg{68.01} \\
		& BAC
		& 51.32 & 70.33 & 92.73 & 94.11 & 60.60 & 62.39 & 82.88 & 73.48 & \dpos{0.86} & 80.91 & \dneg{82.23} \\
		& SpeCa
		& 53.72 & 69.29 & 93.02 & 92.80 & 51.01 & 59.99 & 77.66 & 71.07 & \dpos{2.96} & 65.34 & \dneg{54.69} \\
		
		\bottomrule
	\end{tabular}
\end{table*}

\subsection{Case Study: Action Trajectories Analysis}
\label{sec:case-study}

\Cref{fig:case-study} presents paired execution examples for Joint-WAM, Fast-WAM, and Faster-WAM. In these examples, the vanilla policies fail to complete the tasks, whereas WAMotor enables successful execution. For Joint-WAM, the trajectories largely overlap during the initial approach but diverge during subsequent interaction. For Fast-WAM, a slight gripper deviation causes the vanilla policy to get stuck against the upper part of the microwave, whereas WAMotor avoids this deviation. For Faster-WAM, repeated local changes in direction prevent the vanilla policy from grasping the target object, while WAMotor produces more coherent motion and completes the grasp in a single attempt. These examples illustrate how changes in action generation can lead to different outcomes during closed-loop execution.

To further investigate this behavior, we compare action entropy between vanilla inference and WAMotor. As shown in \cref{tab:speculative-entropy}, mean entropy decreases from 2.31 to 2.01 bits for Joint-WAM, from 2.15 to 1.80 bits for Fast-WAM, and from 2.09 to 1.87 bits for Faster-WAM. These consistent reductions indicate lower uncertainty in action outputs under the measured entropy metric. Together with the trajectory visualizations, these results suggest that WAMotor can promote more consistent action generation, complementing the task-level results reported in the main experiments.

\subsection{Complementarity with Diffusion Caching}
\label{sec:complementarity}

WAMotor reuses velocities along the denoising trajectory, whereas diffusion caching methods reuse intermediate features inside the network. Since the two operate at different levels, they can be combined. We apply WAMotor on top of VLA-Cache, TeaCache, BAC, and SpeCa, and report the results on LIBERO-Plus in \cref{tab:caching-complementarity}.

Compared with each caching baseline alone (\cref{tab:libero_plus}), adding WAMotor improves both success rate and latency in all 12 model--baseline pairs. Average success rates increase by 0.42--3.28 percentage points, and latency decreases by $1.48\times$--$2.40\times$. Compared with WAMotor alone, the combinations further reduce latency by 4.7\%--17.4\% on Joint-WAM, 6.1\%--28.9\% on Fast-WAM, and 5.4\%--27.7\% on Faster-WAM, while maintaining average success rates close to Vanilla. The fastest combination, SpeCa + WAMotor, reaches 65.05~ms on Fast-WAM, a $6.02\times$ speedup over Vanilla. These results show that WAMotor is complementary to existing caching methods and can be stacked with them to trade a small amount of accuracy for additional speedup.

\end{document}